\documentclass{article}
\usepackage{iclr2027_conference}
\usepackage{fontspec}
\usepackage{fix-cm}
\usepackage{amsmath,amssymb}
\usepackage{pifont}
\usepackage{booktabs,tabularx,makecell,array,multirow}
\usepackage{graphicx,enumitem,xcolor,microtype,xurl}
\usepackage{caption}
\usepackage{wrapfig}
\usepackage{colortbl}
\usepackage{placeins}
\usepackage{float}
\usepackage{tikz}
\usepackage{hyperref}
\usepackage{widetable}
\usepackage{needspace}

\usetikzlibrary{arrows.meta,positioning,fit,calc}
\hypersetup{colorlinks=true,linkcolor=blue!55!black,citecolor=blue!55!black,urlcolor=blue!55!black}
\newcommand{\asyncrbench}{\textsc{Async-RBench}}
\newcommand{\appendixonly}[1]{#1}
\definecolor{oursbg}{HTML}{EEF3F8}
\newcommand{\covered}{\ensuremath{\checkmark}}
\newcommand{\partialcoverage}{%
  \tikz[baseline=-0.25em]{%
    \fill (0,0) -- (90:0.36em)
      arc[start angle=90,end angle=270,radius=0.36em] -- cycle;
    \draw[line width=0.35pt] (0,0) circle[radius=0.36em];
  }%
}
\newcommand{\notcovered}{\ding{55}}
\newcolumntype{Y}{>{\raggedright\arraybackslash}X}

\title{Async-RBench: Benchmarking Asynchronous\\Task Execution in LLM Agents}
\author{\parbox{0.96\textwidth}{\raggedright
\textbf{Detao Zeng, Guilin Ou, Nuo Chen, Yunhao Feng, Hongliang Wang, Qinli Yang\\ Junming Shao\thanks{Corresponding authors.}\;, Zhong Zhang\footnotemark[1]}\\[2pt]
\normalfont University of Electronic Science and Technology of China\\
\texttt{zengdetao8@gmail.com}, \{\texttt{junmshao}, \texttt{zhong.zhang}\}\texttt{@uestc.edu.cn}}}
\iclrfinalcopy
\begin{document}

\maketitle
\lhead{Preprint}

\vspace{-1mm}

\begin{abstract}
Large language model (LLM) agents increasingly execute tasks asynchronously: a main agent continues working while its subagents run, and the task state may change before their results arrive. The main agent must decide which work to update or preserve and which checks to repeat, while avoiding prohibited changes. Final task success alone does not reveal whether these results were handled correctly, and natural asynchronous arrivals are hard to reproduce. We introduce Async-RBench, a benchmark for asynchronous task execution in LLM agents, with 200 executable tasks covering eight scenarios. A scenario controller regulates result delivery using predefined task-state conditions, while an Event Response Contract (ERC) specifies what each response must change, preserve, avoid, and re-verify. The Dynamic Replanning Score (DRS) weights the response process and outcome equally and is reported separately from task success. We evaluate nine agents and find DRS scores of 19.2--45.1, with the lowest scores on duplicate reports and subagent failures. Under our execution protocol, all nine agents obtain lower task success with asynchronous than with batched delivery (26.0\% vs.\ 29.7\% on average). For three agents, prompting the main agent to rebuild its plan after every
result improves DRS by only 2.6--3.5 points. Together, these findings highlight the challenges of reliable asynchronous task execution and the need to evaluate how agents handle incoming results alongside end-to-end task success.\footnote{Data and code are available at: \url{https://pope8zdt.github.io/Async-RBench}}
\end{abstract}


\section{Introduction}
\label{sec:introduction}
Main--subagent architectures are increasingly common in large language model (LLM) agents: a main agent decomposes tasks, delegates subtasks to
subagents, and integrates their results~\citep{magenticone,manageragentgym}. Subagents can run with separate contexts, specialized roles~\citep{metagpt,chatdev,camel}, tools, and instructions~\citep{autogen,agentscope}, and production coding agents such as Claude Code and Codex provide built-in
support for subagents~\citep{claudecodesubagents,codexsubagents}. Rather than idling until every subagent finishes, the main agent can keep working and consume results \emph{asynchronously} as they arrive. We call such an arrival during ongoing execution an \emph{asynchronous interruption}.

\begin{figure}[htpb!]
\centering
\includegraphics[draft=false,width=\linewidth]{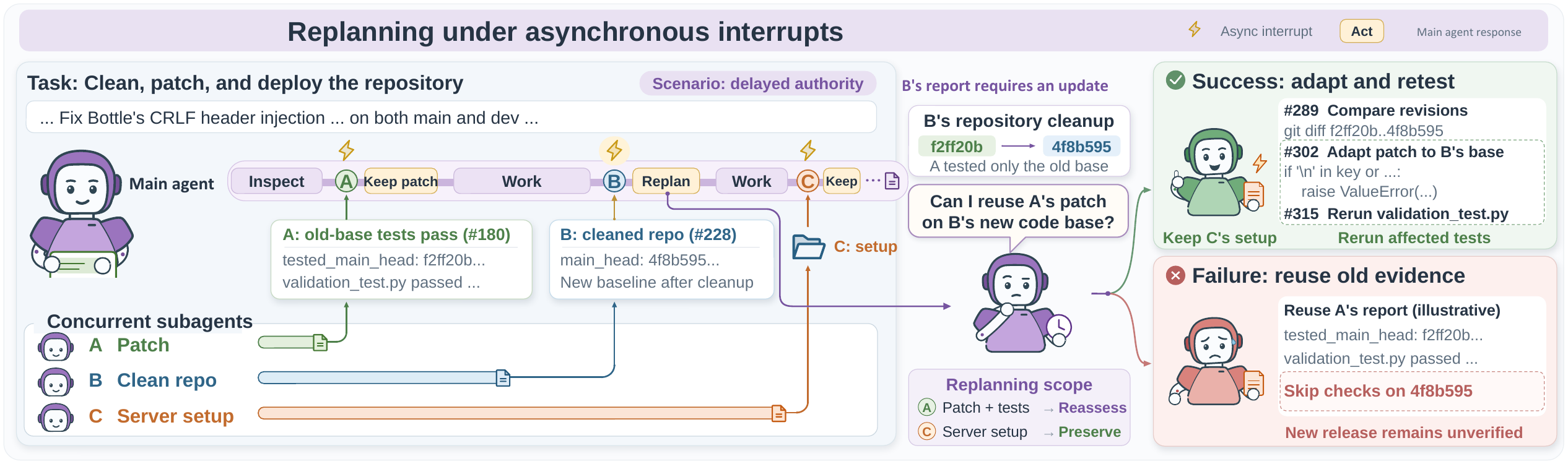}
\caption{A delayed-authority interruption in a repository-repair task. Lightning bolts mark subagent results reaching the main agent, and yellow boxes mark its responses. The right panels contrast the correct response with an illustrative failure.}
\label{fig:execution-modes}
\vspace{-3.5mm}
\end{figure}

The task state may change between a subagent's launch and the arrival of its result~\citep{DBLP:journals/corr/abs-2410-21620,DBLP:journals/corr/abs-2608-18092}. By the time a result arrives, the main agent may already have committed to a plan, written artifacts, or run validation, and the right response depends on how the result relates to that state. A late authoritative result must be acted on and may invalidate earlier work. A stale or duplicate report may require leaving valid work unchanged and avoiding repeated side effects. If a subagent fails, the main agent must recover the affected subtask while preserving valid work on other subtasks~\citep{DBLP:journals/corr/abs-2511-03094}. In Figure~\ref{fig:execution-modes}, subagent B's cleanup changes the code base after subagent A has reported passing tests, so A's report no longer verifies the repository, although no message flags it as outdated. Responding correctly means adapting the patch to the new base, keeping C's server setup, and rerunning the affected tests; reusing A's report leaves the new release unverified. A correct response makes the changes required by a subagent result, preserves valid progress, avoids prohibited side effects, and performs the required validation. Correctness requires both making necessary changes and avoiding unnecessary ones. These errors are not symmetric in cost: ignoring a result that invalidates earlier work leaves artifacts the agent's own evidence no longer supports, whereas rebuilding work the result does not affect repeats side effects, discards verified artifacts, and spends calls the remaining task still needs.

Existing benchmarks do not isolate this capability. Evaluations of interruptible, dynamic, and multi-agent systems inject disruptions through user instructions \citep{interruptbench,agentchangebench,lrminterruptible}, environment events \citep{gaia2,clawarena}, interrupted execution \citep{continuitybench,handoffdebt}, or tool and delegation failures \citep{toolmaze,orchestrabench}, most of which make explicit that something must be handled. A subagent result may not explicitly indicate that earlier work has become invalid. The main agent must assess the result's validity, recency, and scope to decide whether existing work needs to change. The closest benchmark includes subagent reports but neither specifies per-event response requirements nor separates event response from task success \citep{clawarenateam}. Evaluation poses two further obstacles. First, final task success alone cannot distinguish correct event handling from success or failure elsewhere in the task. An agent may handle an event correctly yet fail another requirement, or pass the final checks despite repeating a side effect. Second, asynchronous execution is nondeterministic. Subagent results may reach the main agent unpredictably, making responses difficult to compare across runs and agents.

We introduce \textsc{Async-RBench}, a benchmark for asynchronous task execution in LLM agents, with 200 executable tasks adapted from Terminal-Bench, Gaia2, OSWorld, SWE-bench, and MultiAgentBench \citep{terminalbench2,gaia2,osworld,swebench,multiagentbench}. The tasks evenly cover eight interruption scenarios, ranging from delayed authoritative results and stale updates to duplicate reports and subagent failures (Table~\ref{tab:event-themes}). Subagents execute concurrently in isolated workspaces. A scenario controller buffers results and applies case-specific release conditions, with a maximum-hold fallback to avoid deadlock. Therefore, each task's target event reaches the main agent under designed preconditions rather than natural timing (Figure~\ref{fig:paired-flow}). Each target event has an Event Response Contract (ERC) available only to the evaluator. It specifies required changes, valid work to preserve, prohibited side effects, validation evidence, and the expected outcome, without prescribing a unique action sequence. The \emph{Dynamic Replanning Score} (DRS) weights the response process and the event outcome equally and is reported separately from end-to-end task success $S$.

We evaluate nine agents and report three findings: \textbf{(1)~Handling asynchronous results remains challenging.} 
Overall DRS ranges from 19.2 to 45.1, with the lowest scenario means on Duplication and replay (8.6) and Subagent failure (12.2).
\textbf{(2)~Task success is lower under asynchronous delivery.} 
Under our execution protocol, all nine agents have lower task success rate estimates with asynchronous than with batched delivery (26.0\% vs.\ 29.7\% on average), and eight consume more tokens. Selected paired traces illustrate two failure patterns under Async: the main agent terminates before integrating required artifacts, or acts on assumptions before inspecting the underlying task evidence.
\textbf{(3)~Mandatory plan rebuilding yields limited gains.}
Full-rebuild prompts the main agent to rebuild its remaining plan after every subagent result. For the three agents evaluated, it improves DRS by only 2.6--3.5 points while increasing model calls by 7.5--10.2\%.
Together, these findings highlight the difficulty of reliable asynchronous task execution. By evaluating event responses separately from end-to-end task success, Async-RBench provides a basis for comparing agents and assessing improvements in this setting.
\section{Related Work}
\label{sec:related}

\begin{table}[t]
\centering
\caption[Comparison with related benchmarks.]{Comparison with related benchmarks. \emph{Async delivery}: subagent results reach the main agent while it continues executing. \emph{Post-event checks}: behaviors evaluated after an event. \emph{Per-event scoring}: each event response is scored separately from
end-to-end task success. \protect\covered\ explicit; \protect\partialcoverage\ partial or related; \protect\notcovered\ no explicit design.}
\label{tab:related-work-comparison}
\label{tab:related-work-settings}
\label{tab:related-work-capabilities}
\footnotesize
\begin{widetable}{\textwidth}{@{}ll*{7}{c}@{}}
\toprule
\multirow{2}{*}[-2ex]{\normalsize\textbf{Benchmark}}
& \multirow{2}{*}[-2ex]{\normalsize\textbf{Interruption events}}
& {\normalsize\textbf{Setting}}
& \multicolumn{5}{c}{\normalsize\textbf{Post-event checks}}
& {\normalsize\textbf{Eval.}} \\
\cmidrule(lr){3-3}\cmidrule(lr){4-8}\cmidrule(l){9-9}
&
& \makecell{Async\\delivery}
& \makecell{Update\\plan}
& \makecell{Fix\\affected}
& \makecell{Keep\\valid}
& \makecell{No\\side eff.}
& \makecell{Re-\\verify}
& \makecell{Per-event\\scoring} \\
\midrule
Gaia2~\citeyearpar{gaia2} & Environment/tools & \notcovered & \covered & \covered & \partialcoverage & \covered & \notcovered & \partialcoverage \\
Robotouille~\citeyearpar{robotouille} & Delays/state changes & \notcovered & \notcovered & \partialcoverage & \notcovered & \partialcoverage & \notcovered & \covered \\
MultiAgentBench~\citeyearpar{multiagentbench} & Cooperation/competition & \notcovered & \notcovered & \notcovered & \notcovered & \notcovered & \notcovered & \partialcoverage \\
InterruptBench~\citeyearpar{interruptbench} & Requirement changes & \notcovered & \partialcoverage & \partialcoverage & \partialcoverage & \notcovered & \partialcoverage & \notcovered \\
ToolMaze~\citeyearpar{toolmaze} & Tool failures & \notcovered & \partialcoverage & \partialcoverage & \notcovered & \notcovered & \notcovered & \covered \\
OrchestraBench~\citeyearpar{orchestrabench} & Delegation failures & \notcovered & \partialcoverage & \partialcoverage & \partialcoverage & \partialcoverage & \notcovered & \covered \\
ClawArena-Team~\citeyearpar{clawarenateam} & Workspace updates & \covered & \partialcoverage & \partialcoverage & \partialcoverage & \partialcoverage & \partialcoverage & \notcovered \\
\midrule
\rowcolor{oursbg}
\textbf{Async-RBench (ours)} & \textbf{Subagent results} & \covered & \covered & \covered & \covered & \covered & \covered & \covered \\
\bottomrule
\end{widetable}
\end{table}

\paragraph{Agent benchmarks.}
Many agent benchmarks pose a task and emphasize task-level outcomes, covering general assistance~\citep{agentbench,gaia}, web and mobile interfaces~\citep{webarena,visualwebarena,workarena,androidworld}, tool use with users~\citep{taubench,tau2bench,appworldul}, and long-context or MCP-based workflows~\citep{agencybench,mcpuniverse}. Multi-agent benchmarks extend this to collaboration and competition among agents~\citep{multiagentbench} and to orchestration plans evaluated in deterministic simulation~\citep{orchbench}. Other benchmarks provide complementary checks. AgentBoard measures intermediate progress~\citep{agentboard}, ToolSandbox checks milestones along the trajectory~\citep{toolsandbox}, and AppWorld checks for unintended state changes~\citep{appworld}, while recent work examines corrupt success, silent failures, and process-level defects~\citep{procedureaware,false_success,procctrlbench}. These checks provide progress, state-integrity, or process-quality measures.A related line of work diagnoses failures after the fact, labelling trajectories by cause and attributing each failure to the step where the run first departs from a valid solution~\citep{masfailures,agentdebug,agentrx,traceelephant}.
 \asyncrbench{} instead scores the response to each target event against an ERC written for that event and reports DRS separately from task success.

\paragraph{Asynchronous execution and interruptions.}
Multi-agent frameworks give separate agents distinct roles and conversational contexts~\citep{metagpt,chatdev,camel,autogen,agentscope}, and Magentic-One uses an orchestrator that plans, tracks progress, and replans to recover from errors~\citep{magenticone}. \citet{manageragentgym} frame such orchestration by a manager agent as a unifying research challenge. Single-agent loops such as ReAct interleave reasoning and actions with observations from the environment~\citep{react}. LLMCompiler dispatches tool calls in parallel once their dependencies are resolved, passes their outputs to dependent calls, and replans by sending intermediate results back to the planner~\citep{llmcompiler}. On the benchmark side, Gaia2 evaluates agents in independently evolving environments under time constraints~\citep{gaia2}, Robotouille studies asynchronous planning under execution delays~\citep{robotouille}, and ClawArena and SentinelBench cover evolving information and long-running monitoring~\citep{clawarena,sentinelbench}. Other work studies user requirement changes~\citep{interruptbench,agentchangebench}, interruptions during reasoning~\citep{lrminterruptible}, paused or handed-over execution~\citep{continuitybench,handoffdebt}, or injected tool and delegation failures~\citep{toolmaze,orchestrabench}, and HANDRAISER learns when an agent should interrupt a peer~\citep{learninginterrupt}. AgentChangeBench also reports shift-level recovery metrics~\citep{agentchangebench}. A subagent result need not itself require a plan change. It may be authoritative, stale, duplicated, or partial, and the correct response may be to keep existing work unchanged. The closest benchmark, ClawArena-Team, evaluates how a main agent delegates to a fixed subagent pool, assigns permissions, and integrates results~\citep{clawarenateam}, but it does not define a separate response contract for each subagent-result event. Table~\ref{tab:related-work-comparison} compares their event sources, delivery settings, post-event checks, and use of per-event scoring.

\section{Async-RBench}
\label{sec:benchmark}

\begin{figure}[t]
\centering
\includegraphics[draft=false,width=0.97\linewidth]{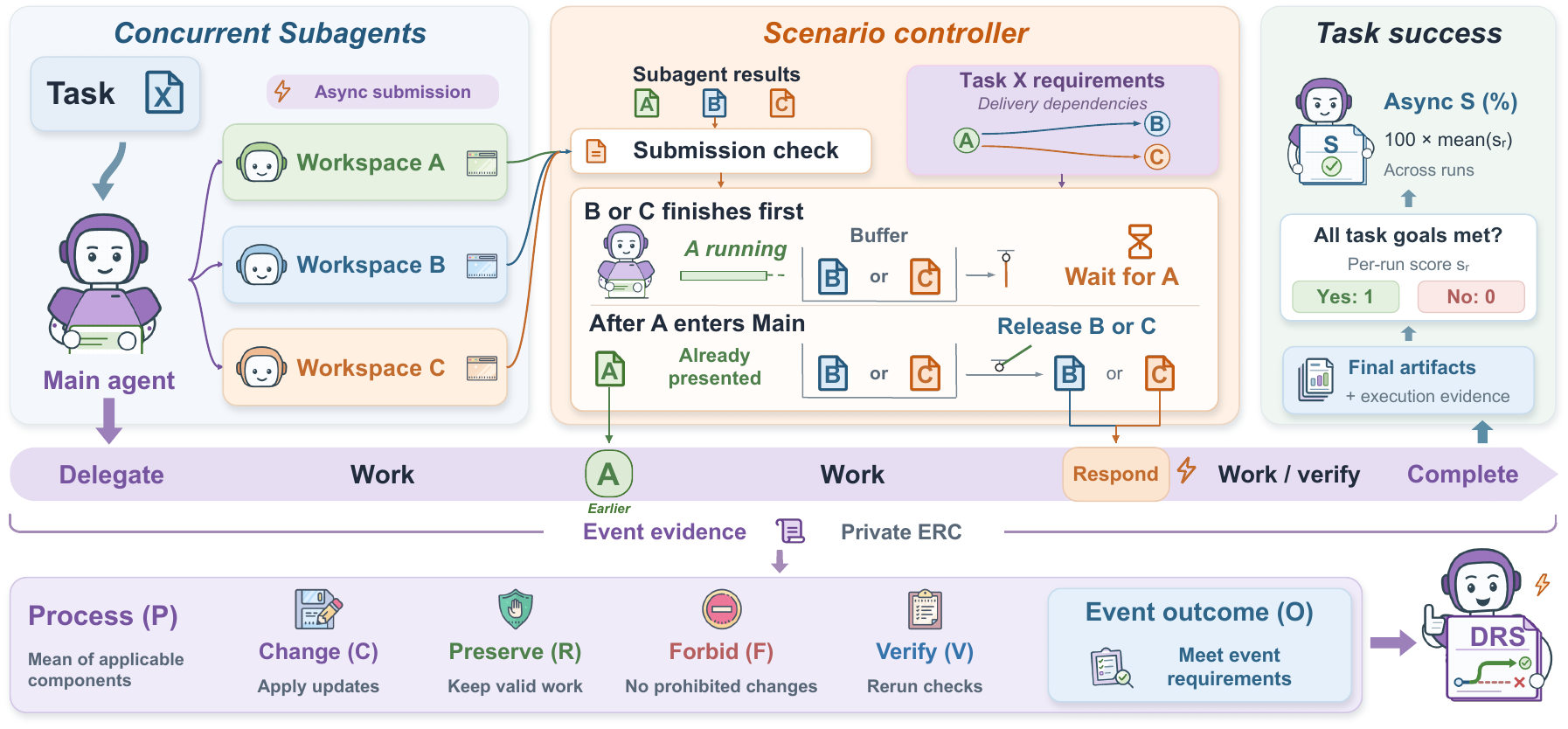}
\caption{Overview of \asyncrbench{}. Subagents A--C run concurrently in separate workspaces while the main agent continues working. The scenario controller checks submitted results and applies case-specific release conditions, with a maximum-hold fallback. B's and C's reports are held until A's is presented, even if they finish first. Task-level checks determine success $S$. The response to the target event and its artifacts are evaluated against the ERC to compute DRS (Section~\ref{sec:metrics}).}
\label{fig:paired-flow}
\end{figure}

\asyncrbench{} addresses both evaluation obstacles identified in Section~\ref{sec:introduction}: controlled delivery applies predefined release conditions (Section~\ref{sec:overview}), and per-event contracts score
the response separately from task success (Section~\ref{sec:metrics}).

\subsection{Controlled Asynchronous Delivery}
\label{sec:overview}

For each task, the main agent delegates subtasks to subagents, which run concurrently in isolated
workspaces while the main agent continues working (Figure~\ref{fig:paired-flow}). Each submitted
result is checked against public requirements on content, files, and format and then held by the
scenario controller. The main agent sees it only once it is \emph{presented}, that is, included in
the input of a main-agent call.

\paragraph{Release conditions.}
Each case specifies the subagent workloads, the relations among their results
(e.g., one result supersedes, duplicates, or conflicts with another), and the
conditions for presenting each result to Main. These conditions depend on
observable task progress, such as which results Main has already received and
which artifacts exist, and are designed so that the prior work referenced by an
ERC is generally available when its event is presented. In
Figure~\ref{fig:paired-flow}, if B or C finishes before A, its result is buffered
until A has been presented in an earlier Main request. Async adds at most one new
result to each Main request. After a result is presented,the main agent receives a
handling window of at least one and, by default, at most four responses;
subsequent results remain queued until this window closes. To prevent indefinite
blocking, the controller releases a result once its maximum hold time is reached
and logs the fallback release (Appendix~C.1).

The controller therefore enforces each event's release prerequisites and
inter-result relations without scripting the exact payload, task state, or
wall-clock timing. It does not tell Main whether to adopt a result, what existing
work it affects, or how to respond. Main must determine the result's validity
and scope, then decide what to revise, what to preserve, and which checks to
rerun.
\paragraph{Event evidence.}
For a presented event $e_i$, we record its result $r_i$, the task state $s_i^-$ immediately before presentation, the main agent's trace $\tau_i$ from presentation to run termination, and the ERC-relevant state $s_i^+$ at termination:
\begin{equation}
Z_i=\left(r_i,s_i^{-},\tau_i,s_i^{+}\right).
\label{eq:event-evidence}
\end{equation}
Presentation logs identify whether the target result reached the main agent. For presented events, $Z_i$ provides the evidence used to score the response.

\subsection{Event Response Contracts and Scoring}
\label{sec:metrics}

\paragraph{Event Response Contract.}
Each target event has an evaluator-only ERC,
\begin{equation}
\Gamma_i=\left(C_i,R_i,F_i,V_i,O_i\right),
\label{eq:event-contract}
\end{equation}
The ERC defines four process components: required changes $C_i$, valid work to preserve $R_i$, prohibited changes $F_i$, and required validation $V_i$. The outcome component $O_i$ specifies the conditions that must hold at run termination. Each component contains scoring points tied to concrete artifacts, operations, or validation records; inapplicable components are excluded from scoring. ERCs specify required state changes, preserved invariants,
prohibited side effects, and validation evidence without prescribing a unique action sequence.

\paragraph{Dynamic Replanning Score (DRS).}
Each component $X_i$ receives a score $\sigma(X_i)\in[0,1]$, the weighted mean of its scoring points.
Process points ($C$, $R$, $F$, $V$) are evaluated on the actions, artifact changes, and validation records
in $\tau_i$, and outcome points ($O$) on the final state $s_i^{+}$. Scoring points, weights, and applicability are fixed before evaluation. The process score $P_i$ averages $\sigma$ over the applicable components among $C_i$,
$R_i$, $F_i$, and $V_i$, with unmet requirements scoring $0$, and the outcome score is $Q_i=\sigma(O_i)$.
For a run with $n$ target events,
\begin{equation}
\mathrm{DRS}=\frac{100}{n}\sum_{i=1}^{n}\frac{P_i+Q_i}{2}.
\label{eq:run-drs}
\end{equation}
Scoring both process and outcome distinguishes cases that task success alone conflates. A run that reaches the required outcome through a prohibited action loses process credit, while a run that fails to reach that outcome can still earn credit for satisfied process requirements. Later repairs can improve the outcome score but do not erase a prohibited side effect.

If the main agent ends the run before the target result
is presented, the event scores $0$. For example, in one Scope and dependencies run, GPT-5.6 Terra adds
late throughput and memory requirements to a research proposal ($\sigma(C)=1$) and keeps the original
research goals ($\sigma(R)=1$) but never substantively checks the revision ($\sigma(V)=0$). No prohibited
change is declared, so $F$ is inapplicable, and the final proposal meets the event requirements ($Q=1$),
giving $\mathrm{DRS}=100\cdot(2/3+1)/2\approx83.3$\appendixonly{ (Appendix~\ref{app:drs-example})}.

\paragraph{Task success and aggregation.}
End-to-end task success $S$ is the percentage of runs whose final artifacts and execution evidence
satisfy all task goals. Each task has one target event ($n=1$). Other subagent messages provide context and
are not scored separately, although later changes to artifacts within the target event's scope are
still checked against its ERC. Scenario DRS averages runs within a scenario, and overall DRS is the
unweighted mean of the eight scenario scores.

\subsection{Tasks and Scenarios}
\label{sec:construction}
\label{sec:scenarios}

\begin{table}[htpb!]
\centering
\caption{Example task and evaluation focus for each scenario, grouped by result validity (first four), information completeness (next two), and dependencies and scheduling (last two). Examples are condensed from actual cases.}
\label{tab:event-themes}
\footnotesize
\definecolor{ScenarioInk}{HTML}{173F5F}

\definecolor{TaskBlue}{HTML}{C9E2DB}
\definecolor{TaskLavender}{HTML}{C9E2DB}
\definecolor{TaskYellow}{HTML}{C9E2DB}
\definecolor{TaskGreen}{HTML}{C9E2DB}

\definecolor{TaskPink}{HTML}{F6D5C8}
\definecolor{TaskMint}{HTML}{F6D5C8}

\definecolor{TaskGray}{HTML}{D9DEE9}
\definecolor{TaskPeach}{HTML}{D9DEE9}
\newcommand{\ScenarioTaskCard}[2]{%
  \tikz{\node[fill=#1, rounded corners=2.5pt, inner xsep=3pt, inner ysep=1.5pt,
    outer sep=0pt, text width=\dimexpr\linewidth-6pt\relax, align=left,
    font=\itshape] {#2};}%
}
\setlength{\tabcolsep}{3pt}
\renewcommand{\tabularxcolumn}[1]{m{#1}}
\begin{tabularx}{\linewidth}{@{}
  >{\raggedright\arraybackslash\color{ScenarioInk}}m{0.17\linewidth}
  >{\hsize=1.1\hsize\linewidth=\hsize\raggedright\arraybackslash}X
  >{\hsize=0.9\hsize\linewidth=\hsize\raggedright\arraybackslash}X@{}}
\toprule
{\color{black}\textbf{Scenario}} & \hspace{3pt}\textbf{Example task} & \textbf{Evaluation focus}\\
\midrule
Delayed authority
& \ScenarioTaskCard{TaskBlue}{Archive an AWS invoice; its fields arrive after the folder and summary table are checked.}
& \textbf{Adopt authoritative evidence}; save the attachment and verify the record.\\ \addlinespace[1.5pt]
Staleness and ordering
& \ScenarioTaskCard{TaskLavender}{Negotiate a cart purchase; an old offer arrives after the new \$42 agreement is saved.}
& \textbf{Reject stale updates}; keep the current agreement over old terms.\\ \addlinespace[1.5pt]
Conflicting results
& \ScenarioTaskCard{TaskYellow}{Diagnose slow DB writes; duplicate-index evidence arrives after causes are ranked.}
& \textbf{Arbitrate by evidence}, not arrival order; revise the root-cause ranking.\\ \addlinespace[1.5pt]
Duplication and replay
& \ScenarioTaskCard{TaskGreen}{Build a scheduling app; a completion message is replayed after its tests are integrated.}
& \textbf{Stay idempotent}; avoid repeated side effects and needless rebuilding.\\ \addlinespace[1.5pt]
Partial to complete
& \ScenarioTaskCard{TaskPink}{Merge two spreadsheets into a CSV; column extractions arrive before the complete merge.}
& \textbf{Separate provisional from complete evidence}; confirm row alignment.\\ \addlinespace[1.5pt]
Subagent failure
& \ScenarioTaskCard{TaskMint}{Coordinate restaurant deliveries; a delivery subagent declines its suborder.}
& \textbf{Recover selectively}; reassign the failed suborder and keep valid routes.\\ \addlinespace[1.5pt]
Scope and dependencies
& \ScenarioTaskCard{TaskGray}{Add slideshow audio; the WAV arrives after a single-slide insertion is planned.}
& \textbf{Extend playback scope}; keep the track and verify playback across slides.\\ \addlinespace[1.5pt]
Resource stragglers
& \ScenarioTaskCard{TaskPeach}{Fix Matplotlib figure pickling; a key diagnosis is delayed as worker slots shrink.}
& \textbf{Prioritize critical work}; weigh waiting against cancelling; keep finished work.\\
\bottomrule
\end{tabularx}
\end{table}

The evaluation set contains 200 tasks adapted from the five source benchmarks (40 each) and divided
evenly among eight scenarios (25 each). The tasks comprise 621 initial subtasks, each assigned to one
subagent\appendixonly{ (Appendix~\ref{app:task-composition})}. A disjoint set of 47 tasks is used only during construction.
Table~\ref{tab:event-themes} groups the scenarios by how a result relates to the current task state:
its validity, its completeness, or its effect on dependencies and scheduling. Each task carries one
primary scenario label, but its ERC, derived from task evidence and dependencies, determines what
must be revised or retained.

\paragraph{Construction and acceptance.}
An LLM agent screens source tasks and traces, with fixed versions and environments, for decision points
at which a delegated result could change an earlier decision. Researchers then check the basis of that
decision, the validity of the new result, and its effect on existing work, design the subagent workflows,
result contents, and release conditions (introducing an event where the source task has none), and derive the ERC from the artifacts expected before the event,
the scope the result affects, and the evidence validation must produce. Evaluated agents never see the
source traces. Before admission, trial runs check that events trigger in the intended state, reference and
equivalent solutions check solvability and that verifiers accept equivalent valid solutions, and invalid variants
(e.g., omitted revisions, incorrect rollbacks, duplicated side effects) check that verifiers reject
them\appendixonly{ (Appendix~\ref{app:construction})}.


\section{Experiments}
\label{sec:experiments}

We first validate DRS (Section~\ref{sec:drs-validity}), then
evaluate agents' responses to asynchronous results (Section~\ref{sec:main-results}), compare task success across delivery modes (Section~\ref{sec:async-effect}), and examine the effect of mandatory plan rebuilding (Section~\ref{sec:full-rebuild}).

\subsection{Experimental Setup}
\label{sec:setup}
\label{sec:e1}

\paragraph{Agents.}
We evaluate Claude Sonnet 5~\citep{claudesonnet5card}, GPT-5.6 Terra and GPT-5.6 Luna~\citep{gpt56card},
Qwen3.7 Plus~\citep{qwen3technical}, Gemini 3 Flash High~\citep{gemini3flashcard},
Qwen3.5 27B~\citep{qwen3technical,qwen35model}, GLM5.3 Flash~\citep{glm5technical,glm53flashmodel},
Kimi K2.5~\citep{kimik25model}, and DeepSeek V4 Pro~\citep{deepseekv3technical}. For each evaluated agent, the main agent and all subagents use the same underlying model. All agents use a common interface, tool set, and evaluation pipeline. We use short names (e.g., Terra, Luna, Gemini) where unambiguous. 

\paragraph{Delivery modes.}
Our main setting, asynchronous delivery (Async), presents results individually while the main agent continues working, following the release rules in Section~\ref{sec:overview}. The baseline, batched delivery (Batched), presents all results together after every required subagent workflow terminates. Both modes run subagents concurrently with the same tasks, workflows, and resource limits, but use different result-delivery protocols. We report DRS for Async and task success $S$ for both modes. Because subagents are executed independently in the two modes, their outputs and the main agent's subsequent actions may differ across runs.

\paragraph{Protocol.}
Each agent is evaluated on all 200 tasks three times per delivery mode, yielding 600 runs per agent per mode. Runs across the two modes are paired by task and repetition index, with matched environments, task and verifier versions, and model configurations. Each run allows at most 100 main-agent responses and 40 responses per subagent execution. Each response is capped at 16,384 output tokens, and at most three subagents run concurrently. A run ends when the main agent calls \texttt{finish}, responds without a tool call, exhausts its budget, or reaches 2,400 seconds. Runs affected by infrastructure failures are rerun\appendixonly{ (Appendix~\ref{app:evidence-boundary})}. For each scenario, we report the mean and sample SD of its three repetition-level mean scores.

\subsection{Validation of DRS}
\label{sec:drs-validity}
\label{sec:validity}

\paragraph{Verifiers agree with human judgments.}
In an initial human audit of 405 randomly sampled runs, automated decisions agree with human judgments on 96.6\% of scored items. All disagreements trace to infrastructure failures, which we fixed before rerunning every affected run\appendixonly{ (Appendix~\ref{app:sampled-review})}.

\begin{table}[t]
\centering
\caption{DRS (0--100) under asynchronous execution. Scenario columns follow Table~\ref{tab:event-themes} and report the mean and sample SD over three repetitions. The Calls column counts main-agent and subagent responses. The last row averages across agents. Bold marks column maxima.}
\label{tab:main-results}
\definecolor{overallbg}{HTML}{EEECFA}
\setlength{\tabcolsep}{1.5pt}
\renewcommand{\tabularxcolumn}[1]{m{#1}}
\newcommand{\drsval}[2]{\makebox[16.5pt][r]{#1}\kern0.5pt{\fontsize{6}{7}\selectfont$\pm$#2}}
\newcommand{\bestdrsval}[2]{\makebox[16.5pt][r]{\textbf{#1}}\kern0.5pt{\fontsize{6}{7}\selectfont$\pm$#2}}
\newcommand{\meanval}[1]{\makebox[16.5pt][r]{#1}\kern0.5pt{\fontsize{6}{7}\selectfont\phantom{$\pm$0.0}}}
\newcommand{\drshead}[1]{{\footnotesize \textbf{#1}}}
{\small
\begin{tabularx}{\linewidth}{@{}l*{8}{>{\centering\arraybackslash}X}>{\centering\arraybackslash}m{26pt}>{\columncolor{overallbg}[\tabcolsep][0pt]\centering\arraybackslash}m{22pt}@{}}
\toprule
\textbf{Agent}
& \drshead{Delay}
& \drshead{Stale}
& \drshead{Conflict}
& \drshead{Replay}
& \drshead{Partial}
& \drshead{Failure}
& \drshead{Scope}
& \drshead{Strag.}
& \drshead{Calls}
& \textbf{DRS}\\
\midrule
Claude Sonnet 5
& \bestdrsval{38.7}{1.8} & \bestdrsval{53.4}{1.2} & \drsval{39.9}{2.1}
& \drsval{15.5}{1.2} & \drsval{29.8}{1.7} & \bestdrsval{35.8}{1.3}
& \drsval{66.9}{1.9} & \drsval{80.5}{1.5}
& 77.1
& \textbf{45.1}\\
GPT-5.6 Terra
& \drsval{34.2}{1.5} & \drsval{43.9}{2.9} & \bestdrsval{41.5}{1.3}
& \bestdrsval{18.5}{1.6} & \drsval{21.4}{1.3} & \drsval{16.2}{1.9}
& \bestdrsval{69.4}{2.0} & \drsval{72.2}{2.4}
& 48.1
& 39.7\\
Qwen3.7 Plus
& \drsval{33.0}{1.3} & \drsval{45.5}{2.9} & \drsval{38.8}{1.5}
& \drsval{12.6}{1.6} & \drsval{21.2}{2.4} & \drsval{6.9}{1.5}
& \drsval{61.8}{2.4} & \bestdrsval{85.9}{2.6}
& 75.9
& 38.2\\
GLM5.3 Flash
& \drsval{29.0}{2.7} & \drsval{47.8}{2.2} & \drsval{27.0}{2.3}
& \drsval{11.6}{2.2} & \bestdrsval{31.2}{2.1} & \drsval{5.0}{0.6}
& \drsval{53.5}{2.4} & \drsval{62.3}{1.4}
& 76.5
& 33.4\\
GPT-5.6 Luna
& \drsval{34.0}{1.2} & \drsval{47.8}{1.2} & \drsval{31.7}{1.4}
& \drsval{4.2}{0.3} & \drsval{22.2}{1.2} & \drsval{14.9}{1.2}
& \drsval{43.8}{1.3} & \drsval{50.5}{2.2}
& 45.5
& 31.1\\
DeepSeek V4 Pro
& \drsval{16.2}{1.6} & \drsval{47.6}{2.2} & \drsval{20.8}{3.2}
& \drsval{2.7}{0.7} & \drsval{24.2}{2.4} & \drsval{16.8}{2.8}
& \drsval{62.0}{2.1} & \drsval{54.7}{3.2}
& 108.0
& 30.6\\
Qwen3.5 27B
& \drsval{28.7}{1.4} & \drsval{32.0}{2.9} & \drsval{31.5}{3.1}
& \drsval{6.1}{0.2} & \drsval{20.0}{1.0} & \drsval{4.2}{0.2}
& \drsval{61.1}{1.4} & \drsval{54.6}{2.2}
& 119.6
& 29.8\\
Kimi K2.5
& \drsval{27.5}{2.3} & \drsval{30.0}{1.9} & \drsval{20.4}{1.3}
& \drsval{3.1}{0.4} & \drsval{12.9}{1.3} & \drsval{8.5}{1.3}
& \drsval{51.4}{1.2} & \drsval{55.3}{2.3}
& 125.2
& 26.1\\
Gemini 3 Flash-H
& \drsval{20.7}{1.1} & \drsval{13.8}{0.4} & \drsval{29.3}{1.3}
& \drsval{2.7}{0.4} & \drsval{8.5}{1.1} & \drsval{1.7}{0.4}
& \drsval{47.2}{2.4} & \drsval{29.9}{2.3}
& 95.0
& 19.2\\
\midrule
Mean
& \meanval{29.1} & \meanval{40.2} & \meanval{31.2}
& \meanval{8.6} & \meanval{21.3} & \meanval{12.2}
& \meanval{57.5} & \meanval{60.7}
& 85.7
& 32.6\\
\bottomrule
\end{tabularx}%
}
\end{table}

\begin{figure}[t]
\centering
\begin{minipage}[t]{0.47\linewidth}
\centering
\includegraphics[draft=false,width=\linewidth]{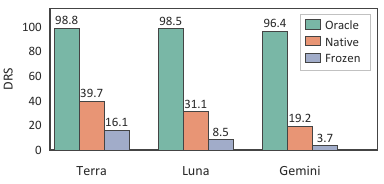}
\caption{DRS of Oracle, Native, and Frozen responses to the same target events.}
\label{fig:drs-validity}
\end{minipage}\hfill
\begin{minipage}[t]{0.5\linewidth}
\centering
\includegraphics[draft=false,width=\linewidth]{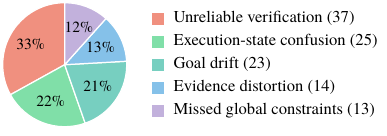}
\caption{Distribution of 112 issue labels across 106 Async runs; a run may receive multiple labels.}
\label{fig:error-category-composition}
\end{minipage}
\end{figure}

\paragraph{DRS separates responses of different quality.}
For Terra, Luna, and Gemini, which rank high, middle, and low in Table~\ref{tab:main-results}, we score three conditions from the same pre-event state (artifacts, main-agent context, subagent states, and event content) under the same limits, ERCs, and scoring rules. \emph{Native} lets the main agent respond freely, as in the main experiment. After the target result is presented, \emph{Frozen} replays a continuation generated by the same model without access to that result. \emph{Oracle} executes a reference recovery program written for that event. As shown in Figure~\ref{fig:drs-validity}, for all three agents, DRS is lowest for \emph{Frozen} (3.7--16.1), higher for \emph{Native} (19.2--39.7), and highest for \emph{Oracle} (96.4--98.8). This ordering shows that DRS distinguishes responses informed by the target result from continuations generated without it. The near-ceiling Oracle scores show that the evaluated ERCs can be satisfied.  Oracle knows each event's requirements, so it is a reference rather than a capability-matched baseline. Frozen earns some credit because unadapted actions can still preserve valid work or avoid prohibited side effects.

\subsection{Agent Performance under Asynchronous Delivery}
\label{sec:main-results}

Overall DRS ranges from 19.2 for Gemini 3 Flash High to 45.1 for Claude Sonnet 5
(Table~\ref{tab:main-results}). Even the highest DRS is below 50 on the 0--100 scale, whereas Oracle range from 96.4 to 98.8 (Figure~\ref{fig:drs-validity}).

\textbf{Scores are lowest on duplicate reports and subagent failures.}
Averaged across agents, DRS is 8.6 on Duplication and replay and 12.2 on Subagent failure, compared with 57.5 on Scope and dependencies and 60.7 on Resource stragglers. The ERCs for the two lowest-scoring scenarios emphasize preserving valid work and avoiding unnecessary side effects. A duplicate report adds no new evidence, while a subagent failure requires recovering the affected subtask without disturbing valid work elsewhere. These averages also include zero scores from valid runs in which the main agent terminates before receiving the target result, and the scenarios use different source tasks. The score gaps therefore do not, by themselves, establish that preserving valid work is the main source of difficulty.

\textbf{Rankings shift across scenarios and differ from task success.}
Claude Sonnet 5 leads overall but trails Terra on Duplication and replay (15.5 vs.\ 18.5), and
Qwen3.7 Plus leads on Resource stragglers (85.9) but scores only 6.9 on Subagent failure. DRS is also
complementary to task success: Qwen3.7 Plus ranks third on DRS but seventh on asynchronous $S$
(Table~\ref{tab:s-comparison}), whereas DeepSeek V4 Pro ranks sixth and third. Across agents, more model calls do not necessarily correspond to higher DRS: Terra scores 39.7 with 48.1 calls per run, whereas Kimi K2.5 scores 26.1 with 125.2 calls\appendixonly{ (Appendix~\ref{app:cost-response-quality})}.

\textbf{Where responses fail.}
Following prior failure taxonomies~\citep{masfailures} and trace-based diagnosis~\citep{agentdebug,agentrx,traceelephant}, we label issues in a sample of 106 Async runs (Figure~\ref{fig:error-category-composition}; excerpts in Appendix~\ref{app:failure-analysis}). Unreliable verification is the most common label; one example is a test that prints a failure message but does not signal failure. Other examples include using a deferred result (execution-state confusion), fabricating a missing receipt (goal drift), presenting static code analysis as runtime coverage (evidence distortion), and accepting plans that pass local checks but jointly exceed a task-wide budget (missed global constraints).

\subsection{Task Success Is Lower under Asynchronous Delivery}
\label{sec:async-effect}
\label{sec:state-failures}

\begin{figure}[t]
\centering
\includegraphics[draft=false,width=1.0\linewidth]{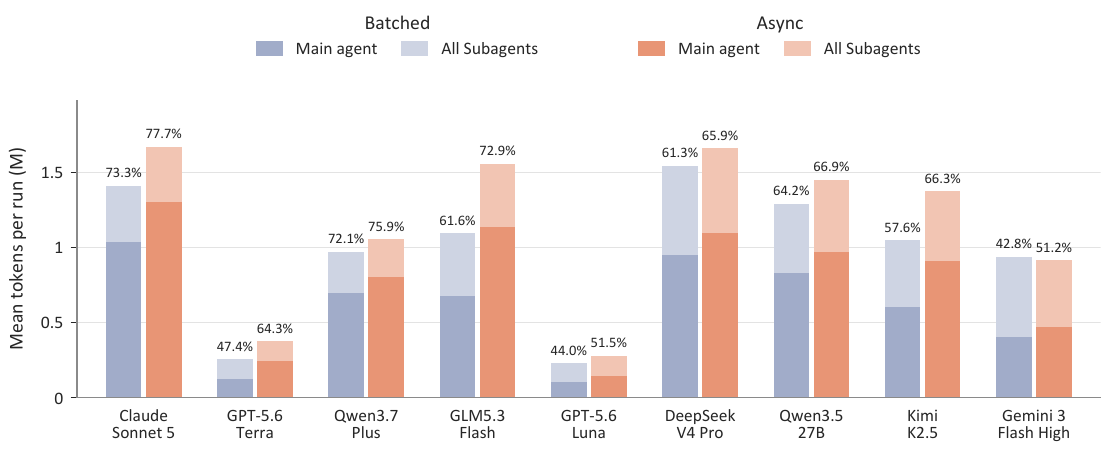}
\caption{Mean token consumption per run by the main agent and all subagents. Each pair shows batched and asynchronous delivery; percentages give the main agent's share.}
\label{fig:cost-main-child}
\end{figure}

\begingroup
\setlength{\columnsep}{10pt}
\setlength{\intextsep}{5pt}
\setlength{\abovecaptionskip}{0pt}
\Needspace{15\baselineskip}

\begin{wraptable}[16]{r}{0.535\textwidth}
\centering
\vspace{-\intextsep}
\caption{Task success rate $S$ (\%) under batched and asynchronous delivery.}
\label{tab:s-comparison}
\vspace{3pt}
\setlength{\tabcolsep}{3.5pt}
\definecolor{DropRed}{HTML}{C0504D}
\newcommand{\drop}[1]{\textcolor{DropRed}{$-#1$}}
\newcommand{\dropbar}[2][DropRed]{%
  \makebox[0pt][l]{\textcolor{black!10}{\rule{30pt}{5pt}}}%
  \textcolor{#1}{\rule{\dimexpr#2pt*5\relax}{5pt}}}
\begin{tabular}{@{}lccrl@{}}
\toprule
\multirow{2}{*}[-2pt]{\textbf{Agent}}
& \multicolumn{2}{c}{\textbf{Task success}} & \multicolumn{2}{c}{\textbf{Change}} \\
\cmidrule(lr){2-3}\cmidrule(l){4-5}
& \textbf{Batched} & \textbf{Async} & \multicolumn{2}{c}{$\boldsymbol{\Delta S}$} \\\midrule
Claude Sonnet 5 & \textbf{42.3} & 38.2          & \drop{4.2} & \dropbar{4.2} \\
GPT-5.6 Terra    & 41.3          & \textbf{38.8} & \drop{2.5} & \dropbar{2.5} \\
Qwen3.7 Plus  & 21.3          & 18.7          & \drop{2.7} & \dropbar{2.7} \\
GLM5.3 Flash   & 25.5          & 20.8          & \drop{4.7} & \dropbar{4.7} \\
GPT-5.6 Luna     & 40.2          & 34.3          & \drop{5.8} & \dropbar{5.8} \\
DeepSeek V4 Pro & 35.7          & 34.7          & \textcolor{DropRed}{$\mathbf{-1.0}$} & \dropbar{1.0} \\
Qwen3.5 27B  & 19.7          & 17.0          & \drop{2.7} & \dropbar{2.7} \\
Kimi K2.5     & 23.2          & 19.2          & \drop{4.0} & \dropbar{4.0} \\
Gemini 3 Flash-H   & 18.3          & 12.3          & \drop{6.0} & \dropbar{6.0} \\
\midrule
\textbf{Mean} & 29.7     & 26.0          & \drop{3.7} & \dropbar[black!45]{3.7} \\
\bottomrule
\end{tabular}
\end{wraptable}

Under our execution protocol, all nine systems obtain lower task-success point estimates with asynchronous than with batched delivery, by 1.0--6.0 percentage points (Table~\ref{tab:s-comparison}), and the mean falls from 29.7\% to 26.0\%. Eight of the nine also consume more tokens, and main-agent consumption rises in all nine (Figure~\ref{fig:cost-main-child}); for Terra it doubles while subagent consumption barely changes\appendixonly{ (Appendix~\ref{app:cost-response-quality})}. Paired traces of tasks solved only under batched delivery illustrate two observed failure paths, without establishing that asynchronous delivery caused them.

\textbf{Ending before required artifacts are integrated.} In a Conflicting results scheduling task, Qwen3.5 27B receives a validation report under asynchronous delivery while the second plan set still sits in another subagent's workspace. The main agent emits its final answer, and shutdown cancels that subagent before it delivers the plans, so the terminal checks find the plan set and the decision manifest missing; the batched run receives both plan sets and passes.

\endgroup

\textbf{Acting before examining the original evidence.}
In a WAL recovery task, two agents that succeed under batched delivery invoke SQLite under asynchronous
delivery before any report arrives and then construct records from assumptions instead of decoding the
encrypted log\appendixonly{ (Appendix~\ref{app:failure-analysis})}.

\subsection{Mandatory Plan Rebuilding Yields Limited Gains}
\label{sec:full-rebuild}

\begin{figure}[t!]
\centering
\includegraphics[draft=false,width=1.0\linewidth,trim=0 12bp 0 0,clip]{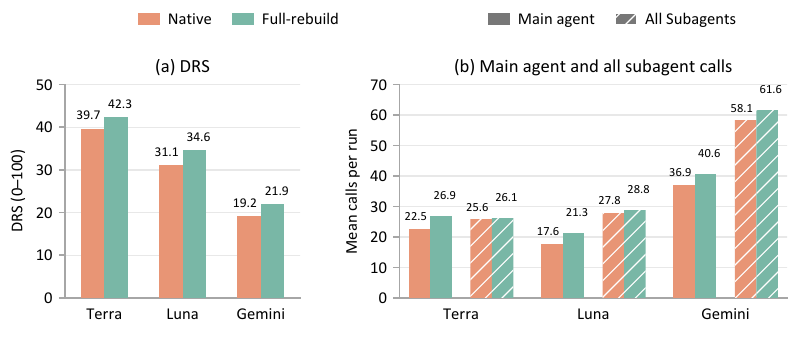}
\caption{DRS and mean model calls per run under Native execution and Full-rebuild. In (b), solid bars count main-agent calls and hatched bars count calls summed over all subagents.}
\label{fig:full-rebuild}
\end{figure}

Under Native execution, the main agent decides for itself whether and how to revise its plan when a
result arrives. Full-rebuild instead instructs the main agent to rebuild the entire remaining
plan from the current task state whenever a subagent result is presented, keeping completed work and
existing artifacts. We run it on Terra, Luna, and Gemini, separately from the main experiment.

Full-rebuild raises DRS on all three agents (Figure~\ref{fig:full-rebuild}a), from 39.7 to 42.3 for Terra, 31.1 to 34.6 for Luna, and 19.2 to 21.9 for Gemini, while mean model calls per run increase by 7.5--10.2\% (Figure~\ref{fig:full-rebuild}b). The additional calls occur mainly in the main agent for Terra and Luna and are distributed roughly evenly between the main agent and subagents for Gemini. Thus, the Full-rebuild instruction yields modest DRS gains at additional computational cost. Because we do not report whether agents actually rebuild their plans as instructed, the results cannot distinguish failures to follow the instruction from errors that persist after rebuilding.
\section{Conclusion}
\label{sec:conclusion-discussion}
\label{sec:conclusion}

We introduced \asyncrbench{} to evaluate how LLM agents handle subagent results asynchronously. Controlled delivery and event-specific contracts allow response quality to be scored separately from final task success. Across nine agents, DRS reaches at most 45.1 and is lowest on duplicate reports and subagent failures. Under our execution protocol, all nine agents have lower task-success point estimates with asynchronous than with batched delivery. For three agents, prompting the main agent to rebuild its plan after every result improves DRS by only 2.6--3.5 points while increasing model calls by 7.5--10.2\%. These findings motivate strategies that track which work a result affects and which remains valid, evaluated through both event responses and end-to-end task success.

\paragraph{Limitations.}
\label{sec:limitations}

Each task isolates one target event, excluding overlaps. DRS conservatively scores valid runs terminating before target delivery as zero, conflating them with response errors. Controller-based delivery with timeout fallback is reproducible but may not reflect real arrivals. Main and subagents share one model, favoring comparability over heterogeneity. Independent subagent runs introduce variation into delivery-timing comparisons. Human-written ERCs and verifiers may omit valid responses despite sampled human review.


\section*{AI Use Statement}

In this work, LLM agents are the object of study: the analyzed trajectories and trained checkpoints are produced by the agents themselves and are reported as experimental results. As authors, we used generative AI tools for tasks requiring disclosure: producing initial annotation labels for trajectory analysis, distilling open-source documentation into the skill library, and assisting with implementing the framework and analysis code. We also used generative AI tools for drafting and polishing text, creating figures, and formatting references. We have reviewed all AI-assisted work: all labels were manually audited, and all code was verified and tested by the authors. We take responsibility for the final content of this work.

\section*{Reproducibility Statement}
Sections \ref{sec:benchmark} and \ref{sec:experiments} elaborate on the full benchmark framework, evaluation metrics, execution modes, model configurations, run budgets, and aggregation procedures. Appendix \ref{app:construction} documents task sources, case construction workflows, scenario specifications, and verifier acceptance criteria; Appendix \ref{app:scenario-examples} presents representative cases for all eight scenarios; Appendix \ref{app:evidence-boundary} specifies result delivery boundaries, termination and rerun rules, cost accounting conventions, and execution and scoring versioning; and Appendix \ref{app:analysis} provides the trajectory evidence underlying the failure analysis. The supplementary material includes the benchmark implementation code, case specification files, verifiers, configuration files, and run manifests, with each analyzed run uniquely tied to its model identifier, configuration parameters, test case, verifier, and protocol version, while private verifier data, access credentials, and license-restricted assets are excluded from the public release package.

\section*{Ethics Statement}
This work introduces Async-RBench, a benchmark for evaluating asynchronous task execution in LLM agents. The study does not involve human subjects, sensitive personal data, or potentially harmful applications. All source tasks are adapted from publicly available, previously published benchmarks (Terminal-Bench, Gaia2, OSWorld, SWE-bench, and MultiAgentBench) in compliance with their respective usage licenses.

This research received no commercial sponsorship, and the authors declare no conflicts of interest. All authors have read and adhere to the ICLR Code of Ethics.

\bibliography{references}
\bibliographystyle{iclr2027_conference}
\clearpage
\appendix
\raggedbottom
\section{Task-Set Construction and Scenario Specification}
\label{app:construction}

\subsection{Task Sources and Construction Process}

Candidate tasks are drawn from
Terminal-Bench, Gaia2, OSWorld, SWE-bench, and MultiAgentBench~\citep{terminalbench2,gaia2,osworld,swebench,multiagentbench},
covering terminal interaction, desktop applications, software engineering, and multi-agent collaboration.
Source tasks provide the original objectives, initial environments, and acceptance criteria, from which we design subagent workflows, result-version relations, and release conditions.

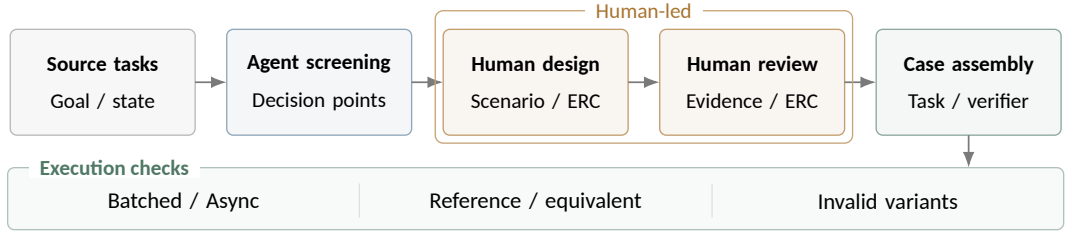
\begin{figure}[!htbp]
\centering
\begingroup
\newfontfamily\constructionfont{Carlito-Regular.ttf}[
  BoldFont=Carlito-Bold.ttf,
  ItalicFont=Carlito-Italic.ttf,
  BoldItalicFont=Carlito-BoldItalic.ttf]
\definecolor{flowblue}{HTML}{456C8B}
\definecolor{flowamber}{HTML}{A87430}
\definecolor{flowgreen}{HTML}{4F7866}
\begin{tikzpicture}[
  x=\linewidth, y=1cm,
  >=Latex,
  stage/.style={
    rounded corners=2.5pt,
    line width=0.5pt,
    draw=black!28,
    fill=black!3,
    anchor=north,
    align=center,
    text width=0.160\linewidth,
    minimum height=14mm,
    inner xsep=3pt,
    inner ysep=4pt,
    outer sep=0pt,
    font=\sffamily\constructionfont\fontsize{8}{10}\selectfont
  },
  flow/.style={->, line width=0.65pt, draw=black!52},
  human/.style={stage, draw=flowamber!70, fill=flowamber!7},
  check/.style={
    align=center,
    text width=0.30\linewidth,
    inner sep=0pt,
    font=\sffamily\constructionfont\fontsize{8.5}{10.5}\selectfont
  }
]
\node[stage] (source) at (0.09,0) {
  \mbox{\textbf{Source tasks}}\\[4pt]
  {\fontsize{8}{10}\selectfont \mbox{Goal / state}}};
\node[stage, draw=flowblue!60, fill=flowblue!6] (candidate) at (0.295,0) {
  \mbox{\textbf{Agent screening}}\\[4pt]
  {\fontsize{8}{10}\selectfont \mbox{Decision points}}};
\node[human] (design) at (0.50,0) {
  \mbox{\textbf{Human design}}\\[4pt]
  {\fontsize{8}{10}\selectfont \mbox{Scenario / ERC}}};
\node[human] (review) at (0.705,0) {
  \mbox{\textbf{Human review}}\\[4pt]
  {\fontsize{8}{10}\selectfont \mbox{Evidence / ERC}}};
\node[stage, draw=flowgreen!65, fill=flowgreen!6] (instance) at (0.91,0) {
  \mbox{\textbf{Case assembly}}\\[4pt]
  {\fontsize{8}{10}\selectfont \mbox{Task / verifier}}};

\draw[flow] (source.east) -- (candidate.west);
\draw[flow] (candidate.east) -- (design.west);
\draw[flow] (design.east) -- (review.west);
\draw[flow] (review.east) -- (instance.west);
\draw[flowamber!65, rounded corners=3pt, line width=0.5pt]
  ([xshift=-3pt,yshift=7pt]design.north west) rectangle
  ([xshift=3pt,yshift=-3pt]review.south east);
\node[font=\sffamily\constructionfont\fontsize{8}{10}\selectfont, text=flowamber,
  fill=white, inner xsep=4pt, inner ysep=1pt]
  at ([yshift=7pt]$(design.north)!0.5!(review.north)$) {Human-led};

\draw[flowgreen!45, fill=flowgreen!4, rounded corners=2.5pt, line width=0.5pt]
  (0,-1.83) rectangle (1,-2.65);
\node[anchor=west, font=\sffamily\constructionfont\fontsize{8}{10}\selectfont\bfseries,
  text=flowgreen, fill=white, inner xsep=4pt, inner ysep=1pt]
  at (0.02,-1.83) {Execution checks};
\draw[flow] (instance.south) -- (0.91,-1.83);
\draw[flowgreen!25, line width=0.4pt] (0.333,-2.04) -- (0.333,-2.49);
\draw[flowgreen!25, line width=0.4pt] (0.667,-2.04) -- (0.667,-2.49);
\node[check] at (0.1665,-2.28) {\mbox{Batched / Async}};
\node[check] at (0.50,-2.28) {\mbox{Reference / equivalent}};
\node[check] at (0.8335,-2.28) {\mbox{Invalid variants}};
\end{tikzpicture}
\endgroup
\caption{Task construction and acceptance. LLM-agent screening feeds human-led scenario and ERC design and review; assembled instances then undergo execution checks in both delivery modes, with reference or equivalent solutions and invalid variants.}
\label{fig:appendix-construction-pipeline}
\end{figure}

Construction begins by fixing the source version, task identifier, and initial environment, and checking that the task has an explicit objective, an executable workflow, and a verifiable final state.
An LLM agent then flags decision points that subagent results may affect; its suggestions only support screening.
Researchers examine information dependencies to identify work that can be delegated and the plans, artifacts, or checks that new results may affect, and they determine the scenario labels and scoring requirements.
They then design the subagent workflows, result contents, version relations, release conditions, and event timeline, and write the Event Response Contract (ERC).
Human review checks that the source evidence, the validity of the new result, its task consequences, and the scoring requirements agree; a candidate becomes an executable instance only when it is clear which work must be revised, preserved, or revalidated after the event.
Figure~\ref{fig:appendix-construction-pipeline} summarizes the process, and Appendix~\ref{app:case-spec} describes execution acceptance.

\subsection{Task Composition and Aggregation}
\label{app:task-composition}

The construction set contains 47 tasks, and the evaluation set contains 200 tasks, with no overlap between them.
The evaluation set covers eight scenarios with 25 tasks each and five original sources with 40 tasks each. Its 200 tasks comprise 621 initial subtasks: 36.5\% of tasks start with two subagents, 20.0\% with three, and 43.5\% with four or more, and 52.0\% are rated medium and 48.0\% hard.

Evaluation weights tasks equally, without additional scenario weights.
Because all eight scenarios contain the same number of tasks, averaging scenario means is equivalent to averaging all tasks under complete coverage; each scenario contributes one eighth of the overall result.
The construction set is excluded from all reported results.

\subsection{Sampled Human Review and Verifier Acceptance}
\label{app:sampled-review}

Human review during construction checks that task requirements are clear, required inputs are available, and scoring conditions match the task objectives.
Following work that assesses automatic trajectory evaluators against expert judgments~\citep{agentrewardbench}, we also audit the verifiers.
We randomly sampled 405 run records from the evaluation set. Auditors examined task requirements, initial states, presented results, trajectories, tool outputs, and final artifacts; a model's completion claim alone was never taken as evidence of success.
A verifier decision counts as a false rejection if the auditor judges an artifact valid but the verifier rejects it, and as a missed violation in the opposite case.
In the initial audit, item-level agreement between human judgments and automated decisions was 96.6\%.
Every disagreement traced to external infrastructure failures; no task-design issue was found in the sample.
We then screened all run records across models, modes, and repetitions for the same anomaly types, fixed the environment, and reran all affected runs. The reruns passed run-validity checks and replaced the affected records, while task failures attributable to the models were retained.
These conclusions apply to the audited sample and versions; they do not establish that the verifiers have no false rejections or missed violations.

\subsection{Asynchronous Scenario Taxonomy}

The eight scenarios are organized by result validity, information completeness, task dependencies, and execution scheduling.
Each instance has one primary scenario label and may require revision, preservation, avoidance of forbidden changes, and revalidation.
The label describes the principal event relation; task evidence and dependencies determine the specific response requirements.

\paragraph{Result validity.}
In \emph{Delayed authority}, the main agent progresses using provisional information before a result arrives that supersedes or corrects it, requiring revision of the affected work.
In \emph{Staleness and ordering}, an older result is presented after a newer state has taken effect; the main agent must recognize the version relation and avoid rolling back valid state.
In \emph{Conflicting results}, outputs from valid workflows support different conclusions; the main agent must resolve them using provenance, scope, and task constraints.
\emph{Duplication and replay} requires idempotent handling of repeated results, avoiding duplicate writes, repeated submissions, or unnecessary rebuilding.

\paragraph{Information completeness.}
\emph{Partial to complete} presents local evidence or an intermediate artifact before the full result. The main agent must distinguish provisional information from completion evidence and complete integration and validation once the full result arrives.
In \emph{Subagent failure}, a subagent fails to complete its assigned work, reports failure, or declines further execution. The main agent must recover the affected subtask while preserving valid work completed by other subagents.

\paragraph{Task dependencies and execution scheduling.}
\emph{Scope and dependencies} requires the main agent to propagate changes in scope, dependency relations, or acceptance conditions to the relevant artifacts and checks while preserving unrelated work.
\emph{Resource stragglers} combines delayed critical results with changing resource constraints. The main agent must use task dependencies to choose among waiting, cancellation, reassignment, and continued execution while preserving valid completed state.

One event may impose several response requirements. For example, a late authoritative result may require revising one dependent file, preserving unaffected test results, and rerunning the relevant checks on the updated state.
The ERC therefore translates scenario relations into specific revision scope, preservation requirements, and validation evidence.

\subsection{Case Specification and Execution Acceptance}
\label{app:case-spec}

Each instance includes public task information and private evaluation information, as summarized in Table~\ref{tab:case-information-boundary}. The main agent can use task requirements, the initial environment, available tools, and presented subagent results. Scenario labels, expected responses, and internal verifier state are withheld from systems under evaluation.

\begin{table}[!htbp]
\centering
\caption{Case components and information boundaries. Public version and dependency evidence comes from the task or presented results; the evaluator retains the full affected scope, response requirements, and scoring evidence.}
\label{tab:case-information-boundary}
\small
\setlength{\tabcolsep}{5pt}
\renewcommand{\arraystretch}{1.10}
\begin{tabularx}{\linewidth}{@{}
  >{\raggedright\arraybackslash}p{0.25\linewidth}
  >{\raggedright\arraybackslash}p{0.18\linewidth}
  Y@{}}
\toprule
\textbf{Information} & \textbf{Access boundary} & \textbf{Content and use} \\
\midrule
\rowcolor{blue!5}[0pt][0pt]
\multicolumn{3}{@{}l@{}}{\strut\hspace{5pt}\textbf{Information available to the main agent}}\\
Task and initial environment
& Always visible
& Goal, initial state, tools, and permitted operations.\\
Subagent result
& After presentation
& New evidence or artifacts, received when included in a main-agent call.\\
Version and dependency relations
& Partly visible
& Supports judgments of result validity and affected work.\\
\midrule
\rowcolor{black!4}[0pt][0pt]
\multicolumn{3}{@{}l@{}}{\strut\hspace{5pt}\textbf{Evaluator configuration and records}}\\
Result-release conditions
& Hidden from the main agent
& Determine when qualified results may be presented.\\
Event Response Contract (ERC)
& Hidden from the main agent
& Defines revision, preservation, forbidden changes, revalidation, and outcome requirements.\\
Hidden verifier
& Hidden from the main agent
& Checks event responses and final task state.\\
Logs and state snapshots
& Retained by evaluator
& Record result presentation, main-agent actions, and state changes.\\
\bottomrule
\end{tabularx}
\end{table}

The ERC binds required changes, preservation of valid work, forbidden changes, revalidation, and response-outcome conditions to concrete artifacts, states, operations, or validation evidence.
It specifies the required task state without prescribing a unique action sequence. A reference execution provides one feasible path; other implementations that satisfy the same requirements should also pass.

Before admission, each instance undergoes execution acceptance.
We check that the environment starts, subagents can submit qualified results, events trigger in the intended state, and the verifier can assess both the event response and the final state.
Reference executions and equivalent implementations test feasible completion paths and expected acceptance. Invalid variants test whether the verifier detects omitted revisions, incorrect rollback, duplicate side effects, and missing revalidation.

The task, scenario specification, and verifier are stored under the same version; Appendix~\ref{app:evidence-boundary} describes result delivery and execution rules.

\FloatBarrier
\section{Representative Cases across Eight Scenarios}
\label{app:scenario-examples}

This appendix presents one recorded run for each of the eight scenarios. Each figure retains relevant excerpts from the task, subagent results, and the main agent's subsequent actions. E identifies trace events, \texttt{[...]} marks omissions, and dashed boxes highlight the evidence discussed below. In the figures, ``Main agent'' denotes the main agent and ``Subagent'' denotes a subagent. Fields such as \texttt{replayed} and \texttt{replay\_of\_completion\_id} come from the evaluator's delivery records, which the framework logs separately from the result presented to the main agent. The analysis is limited to the records shown: subagent reports and the main agent's final responses describe their execution but do not independently establish task success.

\begingroup
\makeatletter
\newcommand{\appendixcasepanellabel}[2]{%
  \begingroup
  \edef\@currentlabel{\thefigure(#1)}%
  \label{#2}%
  \endgroup
}
\makeatother
\newcommand{\appendixcasepair}[7]{%
  \begin{figure}[!htbp]
    \centering
    \begin{minipage}[t]{0.485\linewidth}
      \vspace{0pt}\centering
      \includegraphics[draft=false,width=\linewidth]{#1}\par
      \vspace{2pt}{\small (a)\par}
    \end{minipage}%
    \hfill%
    \begin{minipage}[t]{0.485\linewidth}
      \vspace{0pt}\centering
      \includegraphics[draft=false,width=\linewidth]{#4}\par
      \vspace{2pt}{\small (b)\par}
    \end{minipage}\par
    \caption{\textbf{(a)} #2 \textbf{(b)} #5}\label{#7}
    \appendixcasepanellabel{a}{#3}
    \appendixcasepanellabel{b}{#6}
  \end{figure}
}

\subsection{Delayed authority: Select a deployment backend}
The task requires choosing a parallel backend, profile version, and world size from the profile applicable to the final deployment. A subagent first returns authoritative profile v2, recommending pipeline parallelism with a world size of two; a later candidate implementation report still cites v1. The main agent then writes v2 into both the deployment configuration and the batch lineage record. Arrival time and profile applicability must be distinguished: a later report does not necessarily provide a more current basis for deployment.

\subsection{Staleness and ordering: Fix Bottle header injection}
After the repository is sanitized and rewritten, a subagent reports new branch revision identifiers. A patch report arrives later but still cites pre-rewrite revisions as its test basis. The main agent subsequently runs a local header check covering carriage returns, line feeds, and null characters; the shown command exits successfully. The ordering can be misleading: a newly received report may rely on an outdated code state. Test results from old revisions cannot directly substitute for checks of the current repository.

\appendixcasepair
{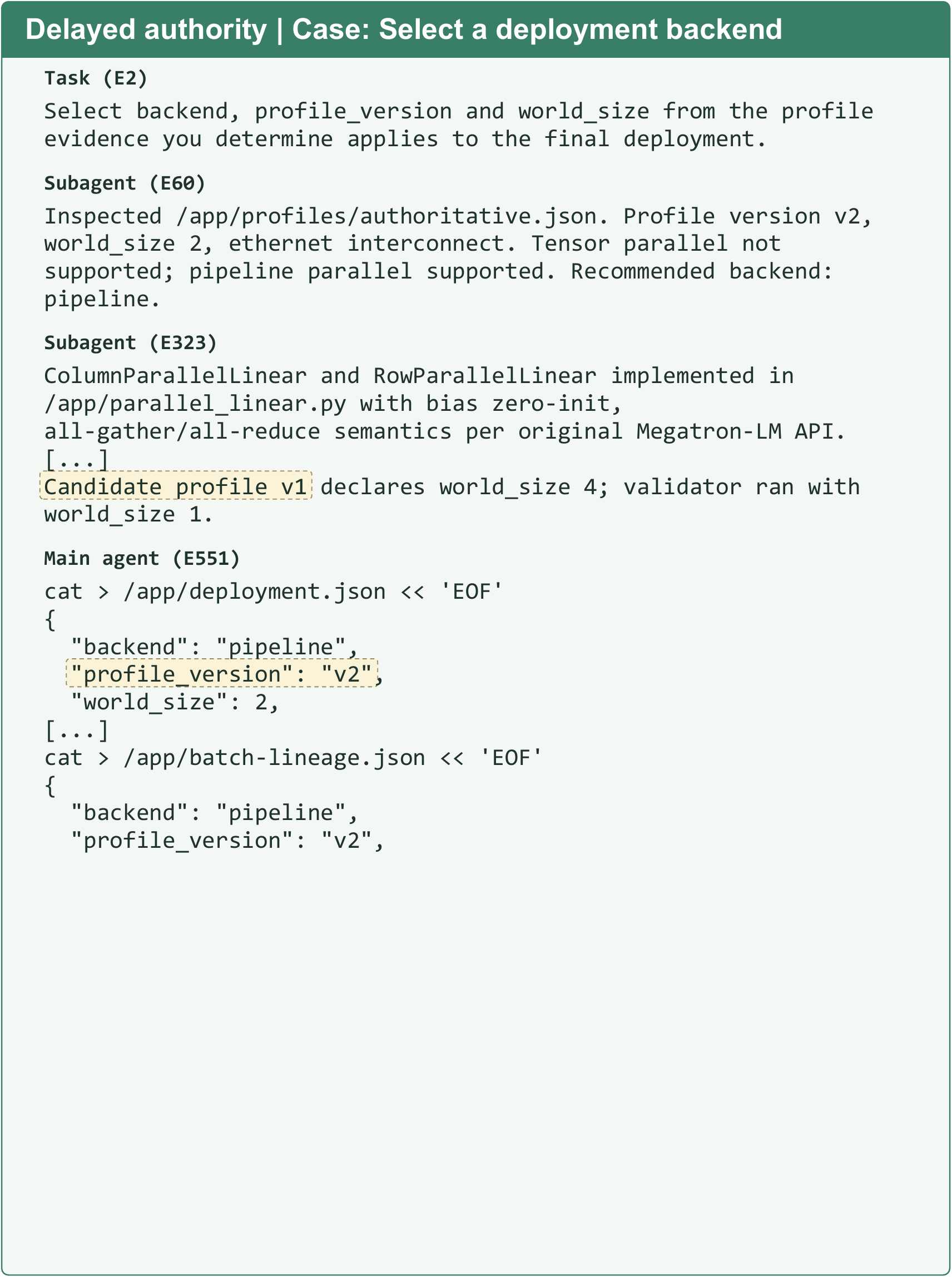}
{Delayed authority: deployment backend selection.}
{fig:case-01-delayed-authority}
{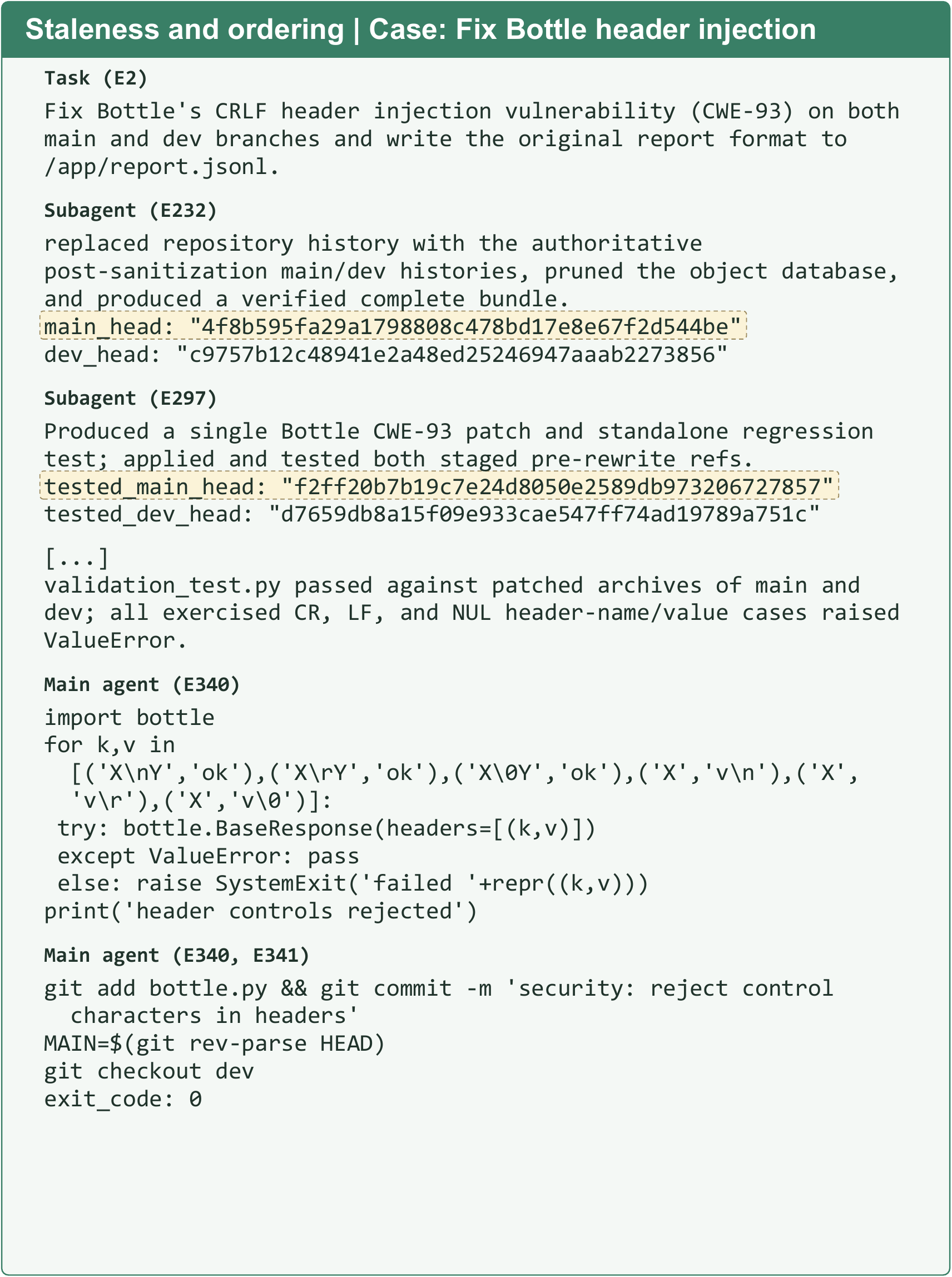}
{Staleness and ordering: Bottle header injection.}
{fig:case-07-staleness-and-ordering}
{fig:case-pair-1}

\subsection{Conflicting results: Recover and remove a Git secret}
Repository scanners return different findings about a secret. Scans of the working tree and ref-reachable history find no match, whereas the object-database report locates the target in an unreachable object. These apparently conflicting results cover different repository scopes. The subsequent recovery uses the object-database finding, and the cleanup report records reflog expiration and garbage collection. The main agent needs to check which repository states each scanner covers rather than choosing between reports solely because they say ``found'' or ``not found.''

\subsection{Duplication and replay: Build OfficeTaskScheduler}
The same completion appears again after its first delivery, with the replay retaining the original completion identifier. The main agent initially rejects the result and later accepts it; this change occurs before the replay. After the replay arrives, the main agent records another acceptance. The figure therefore establishes duplicate delivery and acknowledgment, not that the replay changed the decision or caused duplicate execution. Assessing this response requires checking the completion's identity, the existing decision, and any actual side effects separately.

\appendixcasepair
{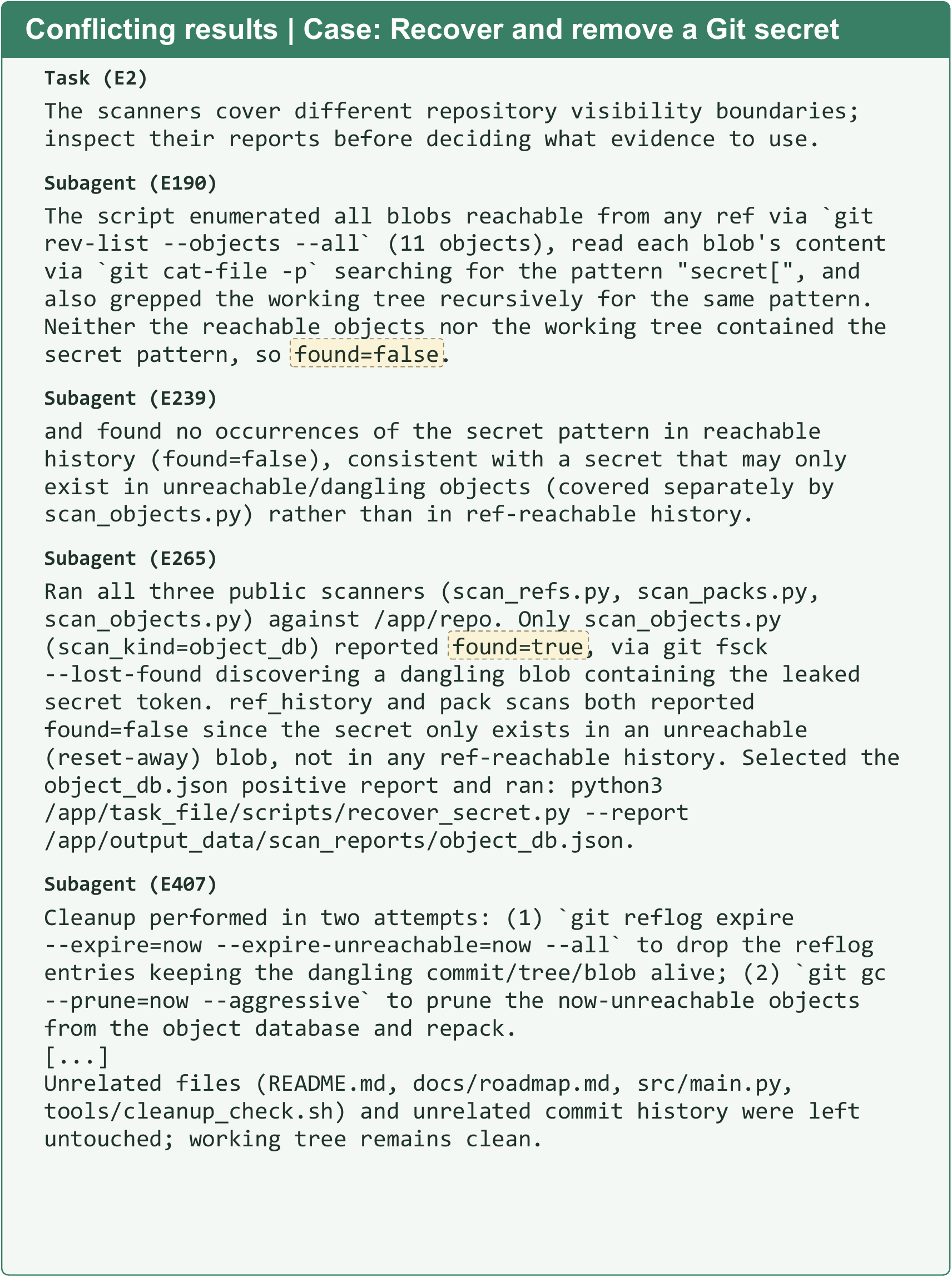}
{Conflicting results: Git secret recovery.}
{fig:case-02-conflicting-results}
{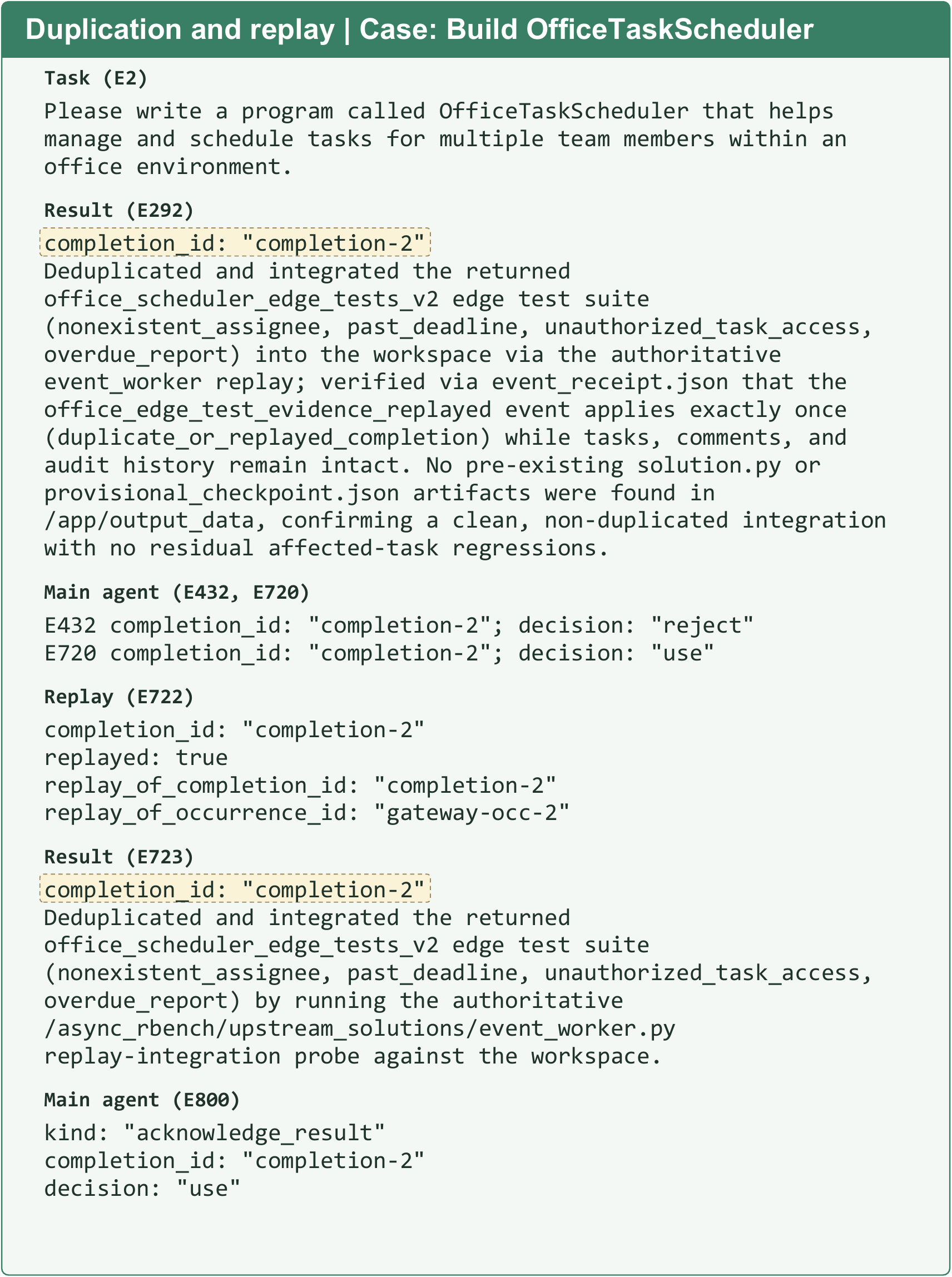}
{Duplication and replay: OfficeTaskScheduler.}
{fig:case-03-duplication-and-replay}
{fig:case-pair-2}

\subsection{Partial to complete: Fix SymPy dimension inference}
The main agent initially adds an exponential-function special case to address the incorrect rejection of a dimensionless exponent. A fuller report then explains that the fix must preserve the meaning of dimensional arguments and check dimensional constraints for other functions. The main agent revises the generic function branch accordingly. A later report, however, warns against extending this normalization to arbitrary user-defined functions. The trace preserves these successive changes in scope: additional information expands the initial repair and then further limits where it should apply.

\subsection{Subagent failure: Build the MultiServe delivery app}
Without the actual application context, a subagent constructs a delivery implementation that groups orders by zone and priority. The main agent rejects this result while accepting results from three other subagents. Its final response attributes the rejection to a mismatch with the actual restaurant, order, and courier interfaces, and states that the main agent implemented the corresponding logistics logic itself. Failure here means an unusable subtask result, not a crashed subagent process. Recovery concerns replacing that result while retaining other completed work.

\appendixcasepair
{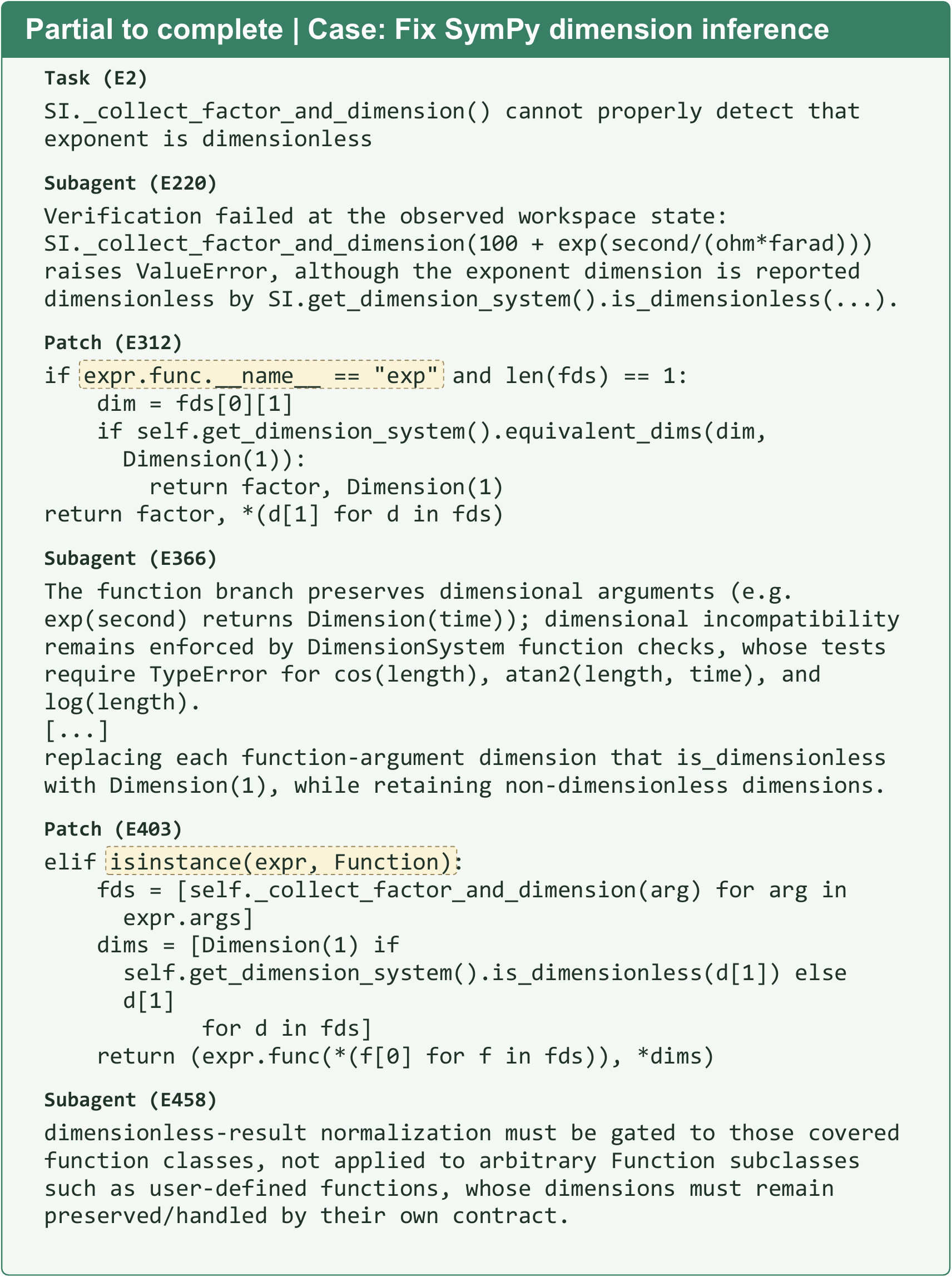}
{Partial to complete: SymPy dimension inference.}
{fig:case-05-partial-to-complete}
{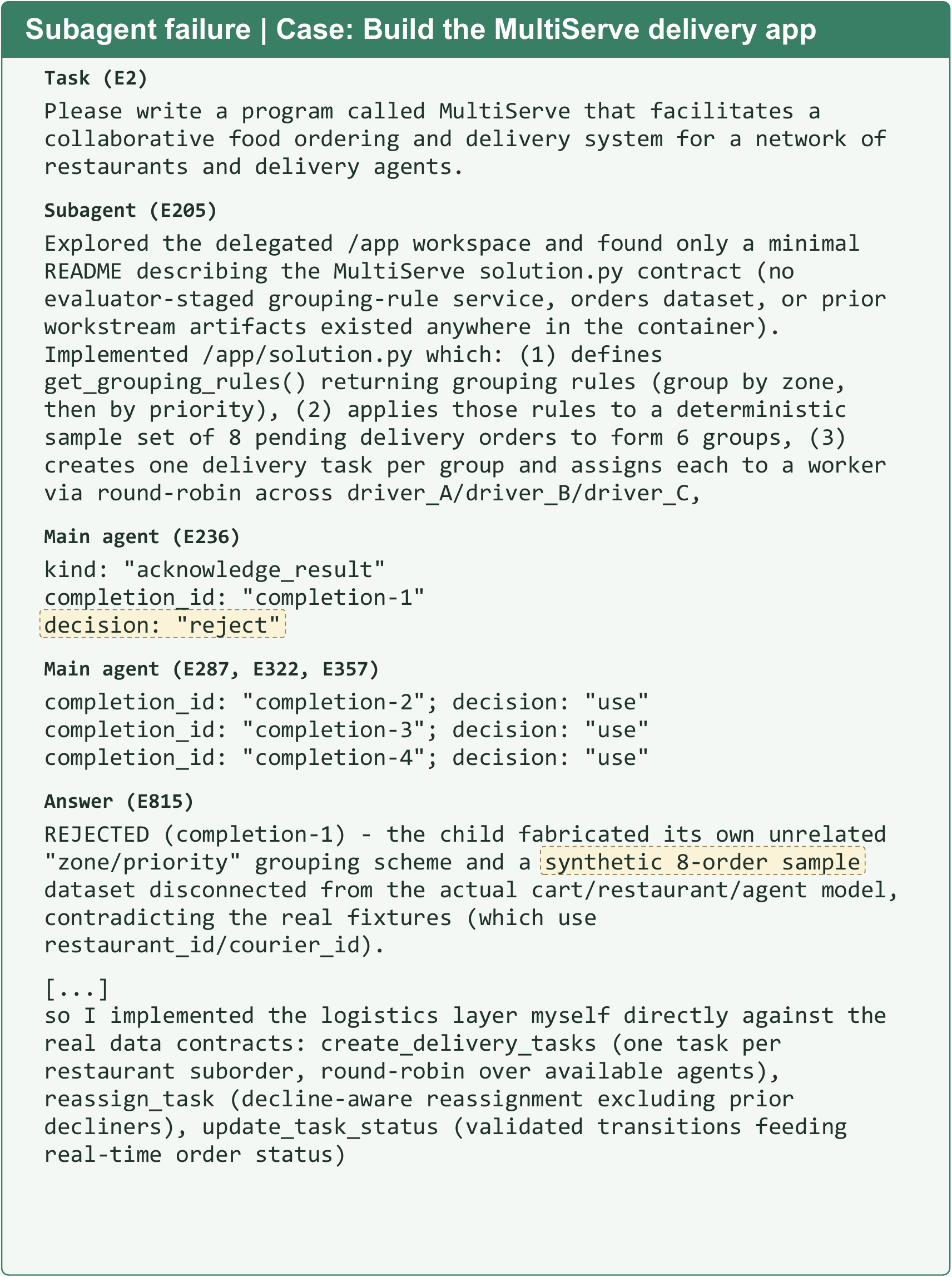}
{Subagent failure: MultiServe delivery app.}
{fig:case-04-child-failure}
{fig:case-pair-3}

\subsection{Scope and dependencies: Fix Pylint regex parsing}
The task addresses commas in regular-expression quantifiers being mistaken for list separators, incorrectly splitting inputs such as \texttt{foo\{1,3\}}. The main agent checks several inputs containing quantifier commas and ordinary separator commas. Subsequent reports examine whether ordinary list parsing and configuration-loading paths are affected and provide corresponding test cases. Because the relevant logic serves multiple configuration entry points, the fix must do more than make the target expression work: it must also preserve the parsing behavior expected by other callers.

\subsection{Resource stragglers: Fix Matplotlib figure pickling}
The case configuration sets the concurrency allowance to one slot; the trace shows a held result followed by a delivery deadline. The returned report states that existing pickling tests pass, but also notes that they do not cover figures after \texttt{align\_labels()} is called. A later targeted check reports unpicklable weak-reference objects, after which the main agent changes how the relevant grouping object saves and restores its state. The important question is what evidence the delayed result adds: passing existing tests does not establish that the task's specific call path works.

\appendixcasepair
{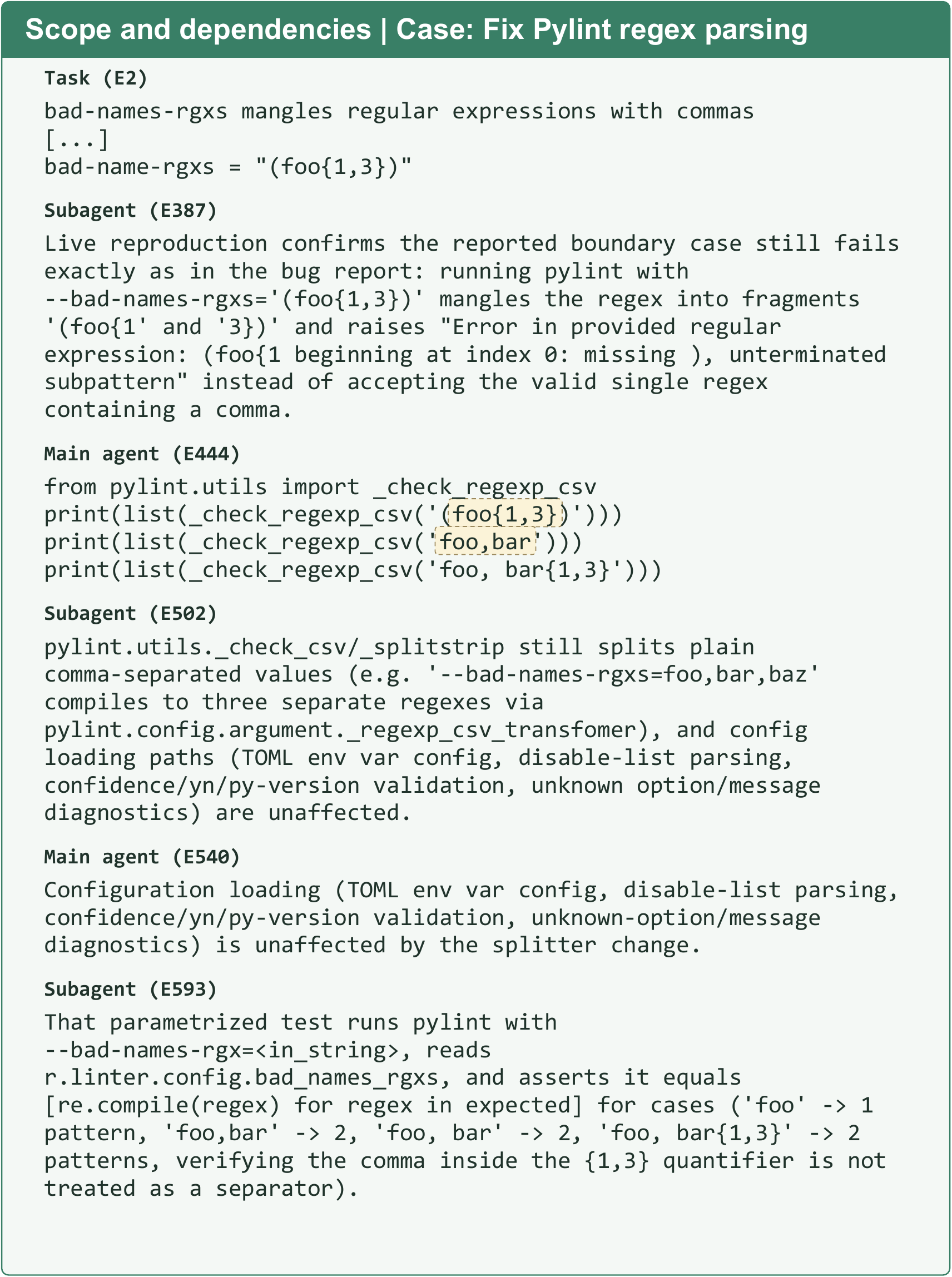}
{Scope and dependencies: Pylint regex parsing.}
{fig:case-06-scope-and-dependencies}
{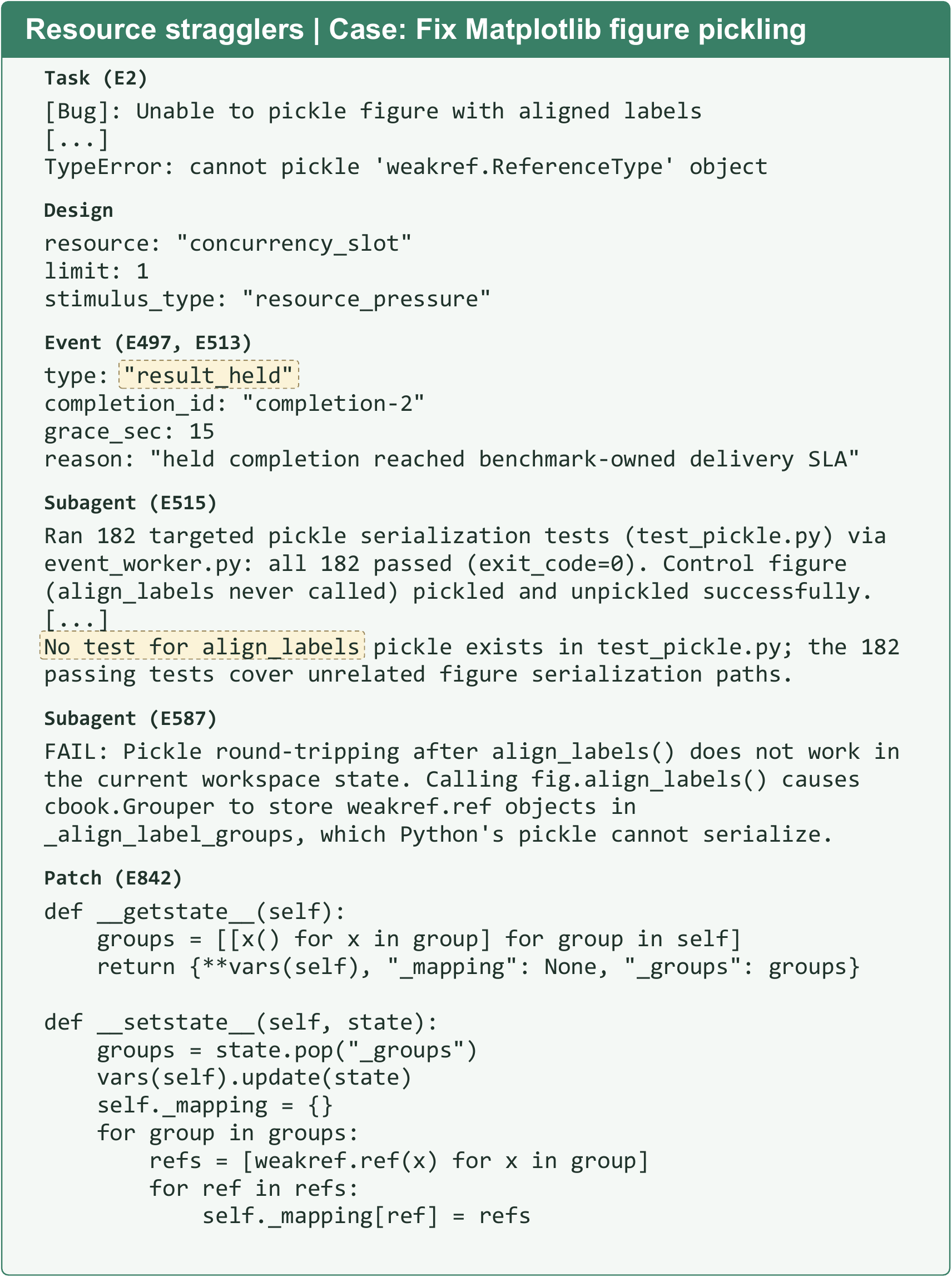}
{Resource stragglers: Matplotlib figure pickling.}
{fig:case-08-resource-stragglers}
{fig:case-pair-4}

\endgroup

\FloatBarrier
\section{Execution Details and Cost Analysis}
\label{app:evidence-boundary}

This appendix describes result delivery, execution limits, run accounting, and cost analysis. Section~\ref{sec:metrics} defines the metrics, and Appendix~\ref{app:scenario-examples} gives examples of the eight scenarios.

\subsection{Result Delivery and Information Boundaries}
\label{app:result-delivery}
Subagent results enter a delivery queue after the result gateway checks their content, files, and structure against public submission requirements.
The framework records the payload and time of each presentation in a main-agent call. This establishes receipt by the main agent; completion or queue entry alone does not.
Replay deliveries are recorded separately.

Both modes execute subagents concurrently. Batched delivery waits for all required workflows to terminate through success, failure, timeout, cancellation, or submission rejection, then presents results together in a stable order.
Asynchronous delivery adds at most one new result per main-agent call while the main agent continues working.
Each result receives a handling window of at least one and, by default, at most four main-agent responses; later deliveries wait until it closes.
This window controls when later results are delivered; closing it does not end event scoring.
If a held result's release conditions are not met within a maximum hold time, the controller releases it anyway and logs the fallback reason; the run in Appendix~\ref{app:drs-example} is such a case.

Scoring uses the state before presentation as a baseline and follows the run through termination.
Process scores use actions, artifact changes, and validation records over this interval;
outcome scores use the relevant saved state at termination.
Repairs after the window closes can therefore satisfy final-state requirements,
while later damage can reduce the outcome score. If a scoring point prohibits a side effect
throughout the response, a subsequent repair does not erase the violation.
The actual termination time is the evidence cutoff, whether the run finishes normally,
stops early, or reaches an execution limit. Evaluators may inspect saved artifacts afterward,
but may not modify them or supply validation that the participant was required to perform.

The workspaces of the main agent and subagents are isolated; subagent artifacts require explicit integration into the main agent's workspace.
Private event roles, affected scopes, verification commands, and answers remain hidden.
Results that pass public submission checks can still be semantically wrong and require the main agent's judgment.
Source trajectories serve only construction and provenance, not as participant inputs, reference answers, or action-matching targets.

\subsection{Execution Budgets, Termination, and Exceptions}
\label{app:execution-limits}
Both modes cap the main agent at 100 responses and each subagent execution at 40. A response counts as one step regardless of its tool calls.
Workflow capacity, recovery delegation, concurrency, and time limits are fixed by configuration.
The main agent and all subagents share a 5,000,000-token emergency cutoff; routine calls do not reserve token quotas.
We record main-agent, subagent, and total usage; equal ceilings need not yield equal costs.

The main agent's first \texttt{finish} call terminates the run. Unpresented results, open response windows, missing submissions, and stale validation are recorded but do not block termination.
Responses without tool calls are labeled as an incomplete main-agent stop or subagent non-submission.
Termination reasons and scores are separate; unscored runs remain available for review.

DRS is computed from the applicable ERC requirements defined in Section~\ref{sec:metrics}.
Prescribed events left unpresented by the main agent's early termination receive zero under that rule
and are not removed from the denominator.
Scorer import and patch-application defects are checked separately rather than attributed directly to the model.
Reruns replace existing positions without increasing the denominator; unreviewed records are not zeros.

Task success is determined by task-level acceptance criteria. A violation
of an ERC-specific process requirement can lower DRS without changing
task success if the task-level criteria are still satisfied.
Token usage of failed runs is retained, and incomplete usage is reported as a lower bound.
Supplementary runs (Frozen, Oracle, Full-rebuild) are reported separately and never replace main-experiment scores.

\subsection{Worked DRS Example}
\label{app:drs-example}

We score one GPT-5.6 Terra run (repetition 1) on a Scope and dependencies case adapted from MultiAgentBench. The task asks the team to write a five-question research proposal on open lookup-free visual tokenization and autoregressive image generation. The case's release conditions place the target result after three other subagent results and a provisional proposal; in this run, the controller's maximum-hold fallback released it before at least one of those results had been delivered. The target result adds codebook-efficiency, throughput, and peak-memory requirements to the proposal's scope. The ERC requires the main agent to integrate the new scope while preserving the original research premises and quality goals. It declares no prohibited change, so $F$ is inapplicable rather than scored as zero. Table~\ref{tab:drs-example} lists the six scoring points.

\begin{table}[!htbp]
\centering
\caption{Scoring points for the worked example. Component scores are weighted means of their points; $F$ is inapplicable.}
\label{tab:drs-example}
\small
\setlength{\tabcolsep}{5pt}
\begin{tabularx}{\linewidth}{@{}llYcc@{}}
\toprule
\textbf{Point} & \textbf{Comp.} & \textbf{Criterion} & \textbf{Weight} & \textbf{Score} \\
\midrule
C1 & $C$ & Incorporate the added efficiency requirements into the final proposal. & 1 & 1 \\
R1 & $R$ & Retain the valid research premises and reconstruction/generation quality goals. & 1 & 1 \\
V1 & $V$ & Substantively check the revised proposal after presentation. & 1 & 0 \\
O1 & $O$ & The final proposal meets the expanded quality and efficiency requirements. & 1/3 & 1 \\
O2 & $O$ & The efficiency evaluation gives concrete measures and comparison conditions. & 1/3 & 1 \\
O3 & $O$ & The presented finding is adopted in the final response, beyond acknowledgment alone. & 1/3 & 1 \\
\bottomrule
\end{tabularx}
\end{table}

The final proposal keeps the reconstruction and generation metrics and adds throughput and peak-memory evaluation, so $\sigma(C)=\sigma(R)=1$ and $Q=1$. Because the task requests a written proposal, describing future experiments is part of the requested artifact and running them is not required; $V$ instead concerns checking the revised proposal, and no substantive post-event check is recorded, so $\sigma(V)=0$. The process score is $P=(1+1+0)/3=2/3$, and $\mathrm{DRS}=100\cdot(2/3+1)/2\approx83.3$. C1 and O1--O3 examine related content from different angles: C1 asks whether the response made the required change, whereas O1--O3 ask whether the final artifact meets the event's requirements, so a run can make a change that still fails the outcome or reach the outcome only through a later repair. V1 was judged from the post-event trace, in which a final assertion or status flag does not count as a check; the trace contains no substantive review of the revised proposal. The example shows why process and outcome are scored separately: a correct final artifact does not establish that the required validation was performed.

\subsection{Run Inclusion, Exclusion, and Reruns}
\label{app:run-inclusion}

Each agent is evaluated on all 200 tasks three times per delivery mode,
giving 600 planned runs per agent per mode, with 75 runs in each scenario.
All runs that can be validly scored are included: failing the task, producing an incorrect result, or exhausting the budget is scored under the normal rules, and low scores never justify exclusion or reruns.
Attempts that cannot be scored because of documented infrastructure faults are rerun with the same task, model, mode, repetition index, and configuration.
Each experimental unit contributes exactly one valid score: reruns replace the attempts they repeat, and the best of several valid attempts is never selected.
Scores are averaged over repetitions within each task and then over tasks with equal weight. Missing scores are never imputed as zero, and any incomplete aggregate reports the numbers of included tasks and runs.
Batched and asynchronous runs are paired by task and repetition index, and paired analyses use only pairs with valid scores in both modes.

\subsection{Cost Composition and Response Quality}
\label{app:cost-response-quality}
\label{app:token-cost-accounting}
\begin{figure}[!htbp]
\centering
\includegraphics[draft=false,width=\linewidth]{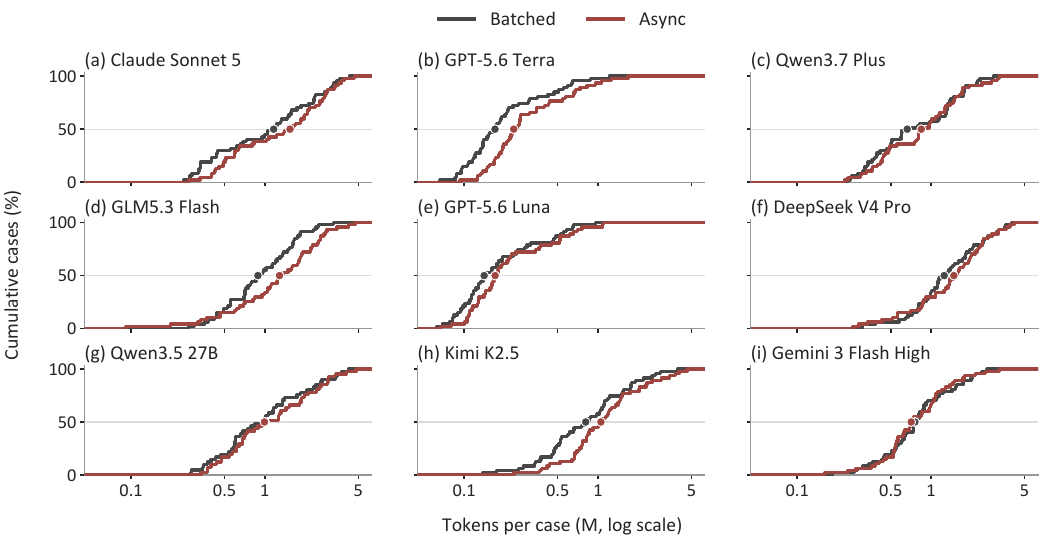}
\caption{Case-level token cost under batched and asynchronous delivery, summed over the main agent and all subagents. Dots mark medians.}
\label{fig:token-cost-cdf}
\end{figure}

We separate main-agent and subagent consumption and compare each system's execution cost with its DRS under Async.
Costs include provider-reported input and output tokens and completed model calls across the main agent and all subagents, retaining reported consumption from failed runs.
We retain paired repetitions with complete cost records for both batched and asynchronous delivery, average repetitions within each case separately for the two modes, and weight cases equally; DRS values come from Table~\ref{tab:main-results}.
The main agent's share is its mean token consumption divided by the mean total token consumption.
The empirical cumulative distributions in Figure~\ref{fig:token-cost-cdf} use the same case means, from which medians and 90th percentiles are also computed.
Missing cost records are not imputed as zero, and unreturned usage is not estimated.

\paragraph{Cost increases span the distribution.}
Eight of the nine agents consume more tokens on average under asynchronous delivery; Gemini 3 Flash High is the exception, with a 2.3\% decrease. Figure~\ref{fig:token-cost-cdf} shows that the increases are not confined to a few costly cases. The medians for Terra and GLM5.3 Flash rise from 0.17M and 0.89M to 0.23M and 1.29M tokens, respectively. The median for Qwen3.5 27B is nearly unchanged, but its 90th percentile rises from 2.67M to 2.97M, so its increase is concentrated in costly cases. A change in mean cost can therefore reflect different changes in the distribution.

\paragraph{Most additional consumption occurs in the main agent.}
Figure~\ref{fig:cost-main-child} shows higher mean main-agent token consumption under asynchronous delivery for all nine agents; the main agent's share of total consumption also increases in every agent.
For Terra, main-agent consumption increases by 100.2\% while subagent consumption is nearly unchanged; for GLM5.3 Flash, main-agent consumption increases by 68.6\% and subagent consumption by only 0.7\%.
Gemini's total consumption falls by 2.3\%, but its main-agent consumption still increases by 17.1\%; the reduction comes from a 16.8\% decrease in subagent consumption.
Changes in total cost can therefore hide different changes in main-agent and subagent consumption, making the main agent an important focus for further analysis of asynchronous execution costs.

\begin{figure}[!htbp]
\centering
\includegraphics[draft=false,width=\linewidth]{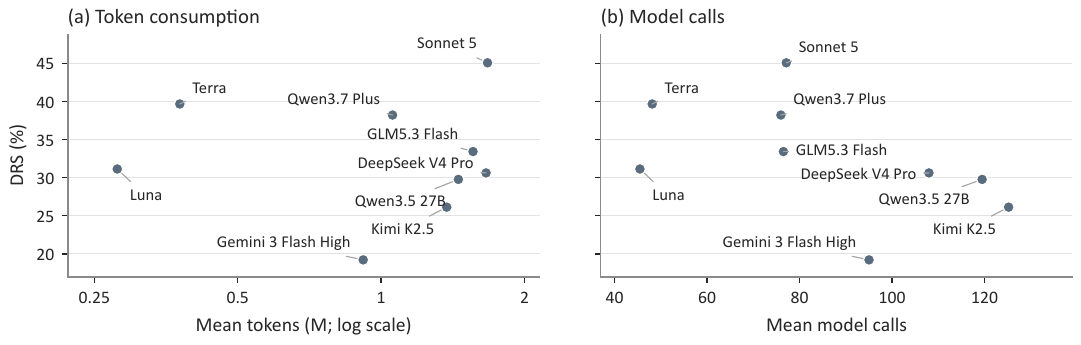}
\caption{DRS versus mean execution costs under Async. Each point represents one agent; costs sum the main agent and all subagents. The token axis uses a log scale.}
\label{fig:cost-drs}
\end{figure}

\paragraph{Similar execution costs correspond to different response quality.}
Figure~\ref{fig:cost-drs} shows that Sonnet and DeepSeek consume 1.67M and 1.66M tokens on average, yet their DRS values are 45.1 and 30.6, respectively.
Model calls show a similar pattern: Sonnet, Qwen3.7 Plus, and GLM5.3 Flash average 75.9--77.1 calls, with DRS values of 45.1, 38.2, and 33.4, respectively.
Execution cost alone therefore does not explain differences in response quality across agents.
These comparisons use average costs and scores from different agents; they do not show how changing the execution budget of a fixed agent would affect its score.

Together, Figures~\ref{fig:cost-main-child} and~\ref{fig:cost-drs} motivate controlled comparisons that hold the model and tasks fixed while varying the main agent's response frequency, context management, and revision strategy, to test whether its execution cost can be reduced while maintaining event response quality.

\FloatBarrier
\subsection{Execution and Scoring Versions}
The manuscript uses the v11.0 execution protocol and the scoring rules defined in Section~\ref{sec:metrics}. Reproduction binds the task set, model and resource configurations, case and verifier versions, execution framework, and individual scoring decisions.

\FloatBarrier
\section{Failure Categories and Trajectory Excerpts}
\label{app:analysis}
\label{app:failure-analysis}
\label{sec:error-analysis}

Figure~\ref{fig:error-category-composition} summarizes the issue labels from 106 Async trajectories of the nine agents. A category is counted at most once per run, and a run may carry several labels, so the 112 labels describe composition rather than the frequency of erroneous operations or their share of DRS loss.

\begin{figure}[!t]
\centering
\includegraphics[draft=false,width=\linewidth,trim=0 0 0 145bp,clip]{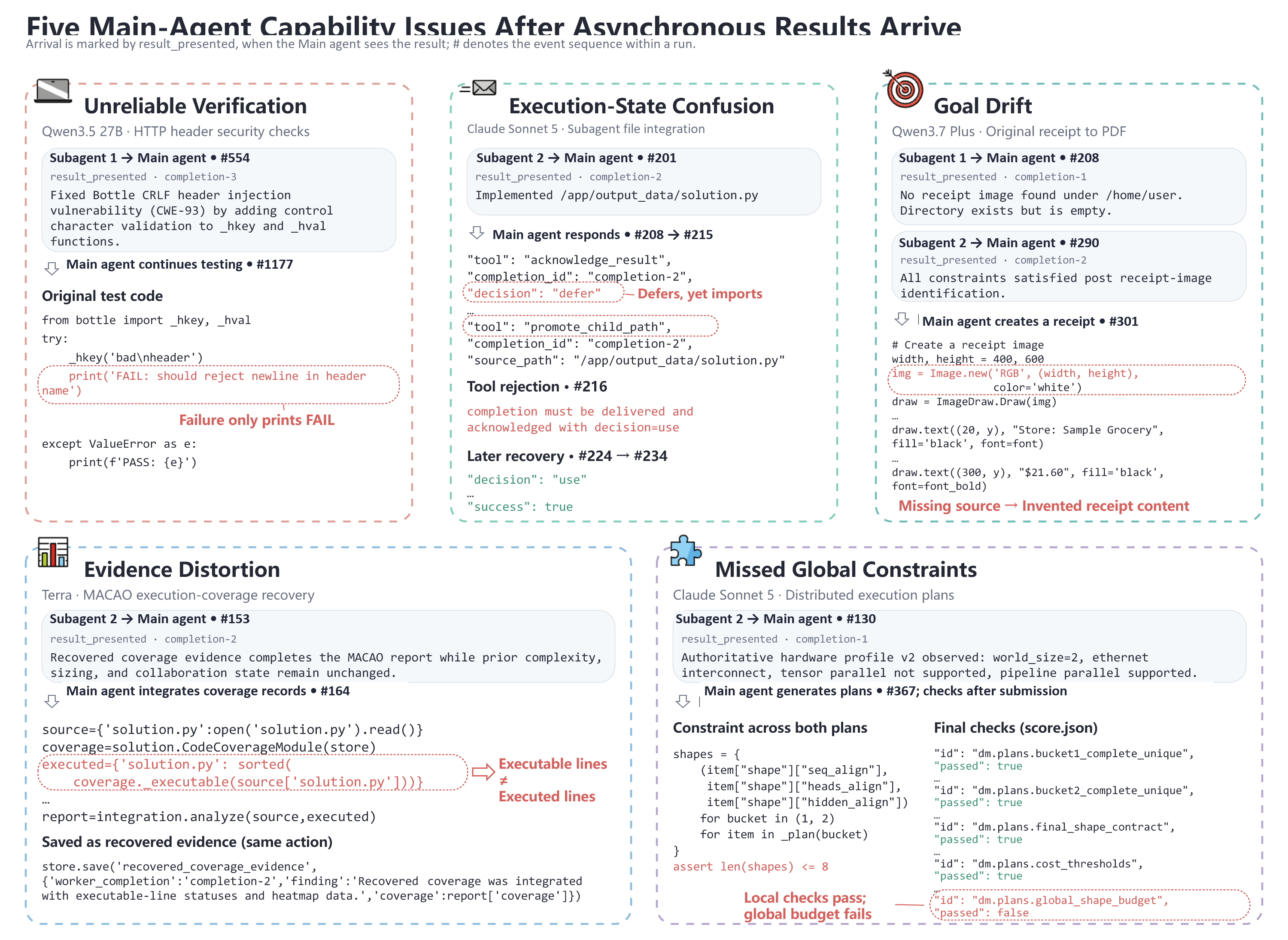}
\caption{Representative issues after asynchronous results arrive. Each panel shows messages, actions, and evidence from an independent run; event numbers give the order. In the panels, ``Main'' denotes the main agent and ``Subagent'' denotes a subagent. The verification example shows a code risk, not an observed execution of the failure branch. Temporal order alone does not show that the interruption caused the error.}
\label{fig:async-error-trajectory-examples}
\end{figure}

Figure~\ref{fig:async-error-trajectory-examples} shows one trajectory excerpt per category, marking when each subagent result is presented and what the main agent does next. A header-injection test prints ``FAIL'' in its failure branch instead of raising an error, so the check never fails (unreliable verification). The main agent defers a result but then tries to import its file, which the tool rejects (execution-state confusion). When no source receipt can be found, the main agent generates a receipt image with invented content (goal drift). The main agent records statically executable lines as executed lines, producing coverage evidence without runtime support (evidence distortion). Two plans pass their local checks but jointly exceed a global shape budget (missed global constraints). These issues do not always cause final failure: the rejected import is later corrected, and the flawed test does not show that its failure branch was ever reached.

\makeatletter\setlength{\@fptop}{0pt}\makeatother

\paragraph{Paired traces.}
In the WAL recovery task of Section~\ref{sec:async-effect}, Claude Sonnet 5 (repetition 1) and Kimi K2.5 (repetition 2) succeed under batched delivery but fail under asynchronous delivery. Both asynchronous runs invoke SQLite before any subagent report arrives, without first backing up or decoding the encrypted log, and then construct records from assumptions: their outputs contain all 11 IDs but leave \texttt{apple.value} at 100 instead of the required 150, failing the data-integrity and log-update checks. The paired batched runs first identify the XOR key \texttt{0x42}, then decode and replay the log. Neither mode's subagent reports supply the full answer, so the difference lies in whether the main agent examines the original evidence before acting on it. In the scheduling task, the paired batched run also completes joint validation of the two plan sets before passing the terminal checks.

\FloatBarrier
\end{document}